%% file: main.tex
\documentclass{article}

\PassOptionsToPackage{numbers, compress}{natbib}

\usepackage[main,final]{neurips_2026}
\newcommand\samethanks[1][\value{footnote}]{\footnotemark[#1]}

\input{setup/package}

\input{setup/macros}

\input{setup/symbols}

\input{setup/graphicspath}

\title{EVIDENT: Routing MLLM Adaptation \\ through Entity-Grounded Visual Evidence \\ for Cross-Domain Video Temporal Grounding}

\author{%
  Geo Ahn\thanks{Equal contribution.} \\
  Kyung Hee University \\
  \texttt{ahngeo11@khu.ac.kr} \\
  \And
    Jiwook Han\samethanks{} \\
  Kyung Hee University \\
  \texttt{mreraser@khu.ac.kr} \\
    \And
    Youngrae Kim\samethanks{} \\
    University of Southern California \\
  \texttt{youngrae@usc.edu} \\
      \AND
      Joonseok Lee \\
    Seoul National University, Hanwha Aerospace \\
  \texttt{joonseok@snu.ac.kr} \\
    \And
    Jinwoo Choi\thanks{Corresponding author.} \\
    Kyung Hee University \\
  \texttt{jinwoochoi@khu.ac.kr} \\
}

\begin{document}

\maketitle

\input{0_abstract}

\input{1_introduction}

\input{2_prior}
\input{3_preliminaries}
\input{4_method}

\input{5_results}
\input{6_conclusion}

\section*{Acknowledgements}
This work was supported by the Institute of Information \& Communications Technology Planning \& Evaluation (IITP) grant funded by the Korea Government (MSIT) under grant RS-2024-00353131 (10\%) and Information \& Communications Technology Planning \& Evaluation (IITP) - Information Technology Research Center (ITRC) grant funded by the Korea government (MSIT) (IITP-2025-RS-2023-00259004, 10\%).
Additionally, it was supported by the National Research Foundation of Korea (NRF) grant funded by the Korea government (MSIT) (RS-2025-02216217; 50\%, RS-2026-25616405; 5\%).
This work was also supported by the MSIT (Ministry of Science and ICT), Korea, under the National Program for Excellence in SW (2023-0-00042, 5\%) supervised by the IITP (Institute of Information \& Communications Technology Planning \& Evaluation) in 2026.

This work was also supported by Youlchon Foundation, National Research Foundation of Korea (NRF) grants, RS-2021-NR05515 (10\%), RS-2024-00336576 (3\%), and RS-2023-0022663 (3\%), and the Institute for Information \& Communication Technology Planning \& Evaluation (IITP) grant, RS-2022-II220264 (3\%). Computing resources were partially supported by a National IT Industry Promotion Agency (NIPA) grant, RS-2026-25621677 (1\%), funded by the Korean government.

{\small
\bibliographystyle{plainnat}
\bibliography{main}
}

\clearpage
\input{9_supple}

\clearpage
\input{checklist}

\end{document}

%% file: setup/package.tex
\usepackage{graphicx}
\usepackage{subcaption}
\usepackage{float}
\usepackage{caption}	%
\usepackage{lscape}                                         %
\usepackage{wrapfig}

\usepackage[lined,ruled,linesnumbered]{algorithm2e}
\usepackage{animate} %

\usepackage{booktabs}                   %
\usepackage{multirow}
\usepackage{makecell}

\usepackage{paralist}
\usepackage{enumitem}

\usepackage{bm}                          %
\usepackage{epsfig}                      %
\usepackage{graphicx}                  %
\usepackage{times}
\usepackage{mathptmx}
\usepackage{mathtools}
\usepackage{amssymb,amsmath}   %

\usepackage{units}
\usepackage{color}
\usepackage[T1]{fontenc}    %
\usepackage{amsfonts}       %
\usepackage[utf8]{inputenc} %
\usepackage{nicefrac}       %
\usepackage{microtype}      %
\usepackage{comment}

\usepackage{url}  %
\usepackage[pagebackref=true,breaklinks=true,letterpaper=true,colorlinks,bookmarks=false]{hyperref}
\usepackage{xspace}
\usepackage[table]{xcolor}
\definecolor{bggray}{gray}{0.9}

\usepackage{setspace}
\usepackage{grfext}
\PrependGraphicsExtensions*{.jpg,.png,.PNG}

\usepackage[most]{tcolorbox}
\usepackage{xcolor}

%% file: setup/macros.tex
\def\eg{e.g.,~}               %

\def\naive{na{\"i}ve\xspace}

\newlength\paramargin
\newlength\figmargin

\newlength\secmargin
\newlength\figcapmargin
\newlength\tabcapmargin

\newcommand{\mpage}[2]
{
\begin{minipage}{#1\linewidth}\centering
#2
\end{minipage}
}

\newcommand{\secref}[1]{Section~\ref{sec:#1}}
\newcommand{\figref}[1]{Figure~\ref{fig:#1}} 
\newcommand{\tabref}[1]{Table~\ref{tab:#1}}

\long\def\ignorethis#1{}

\newbox\jsavebox%

\def\ours{{EVIDENT}}
\def\adapter{{EB Adapter}}
\def\dinoloss{{EB Distillation}}
\def\gating{{E2V gating}}

%% file: setup/symbols.tex
\def\xi{\mathbf{x}_i}

%% file: setup/graphicspath.tex
\graphicspath{{figure}, {example}}

%% file: 0_abstract.tex
\input{figure/fig_teaser}
\begin{abstract}

Fine-tuning MLLMs for Video Temporal Grounding (VTG) often improves in-domain performance but degrades sharply under domain shift. 
In this work, we find that this failure is primarily driven not just by unseen query concepts, but by \textit{visual domain shift}, which prevents the model from coupling its learned temporal localization knowledge with its inherent entity-attention capability.
To address this, we introduce \ours{}, a parameter-efficient adaptation framework that anchors temporal grounding in the inherent entity-attention of pre-trained MLLMs by routing VTG adaptation through \textit{explicit visual entity evidence}.
\ours{} consists of three components: (i) an Entity Bottleneck Adapter that transforms dense visual tokens into compact entity-level slots, (ii) an Entity-Binding Distillation loss that instills objectness priors into the semantically unstructured MLLM visual space, guiding each slot to bind to a coherent entity, and (iii) an Entity-to-eVidence gating mechanism that leverages the captured entities as evidence, steering the model to localize moments containing query-relevant entities.
Together, these components enable VTG fine-tuning to rely on entity-grounded evidence rather than brittle dataset shortcuts.
Experiments on cross-domain VTG benchmarks show that \ours{} consistently improves out-of-domain robustness while preserving competitive in-domain performance with modest parameter overhead. 
These results suggest that entity-level grounding is an effective inductive bias for generalizable temporal localization.
The code is available at \href{https://github.com/KHU-VLL/EVIDENT}{https://github.com/KHU-VLL/EVIDENT}.

\end{abstract}

%% file: figure/fig_teaser.tex
\begin{center}
\centering
\vspace{-2em}
\includegraphics[width=\linewidth]{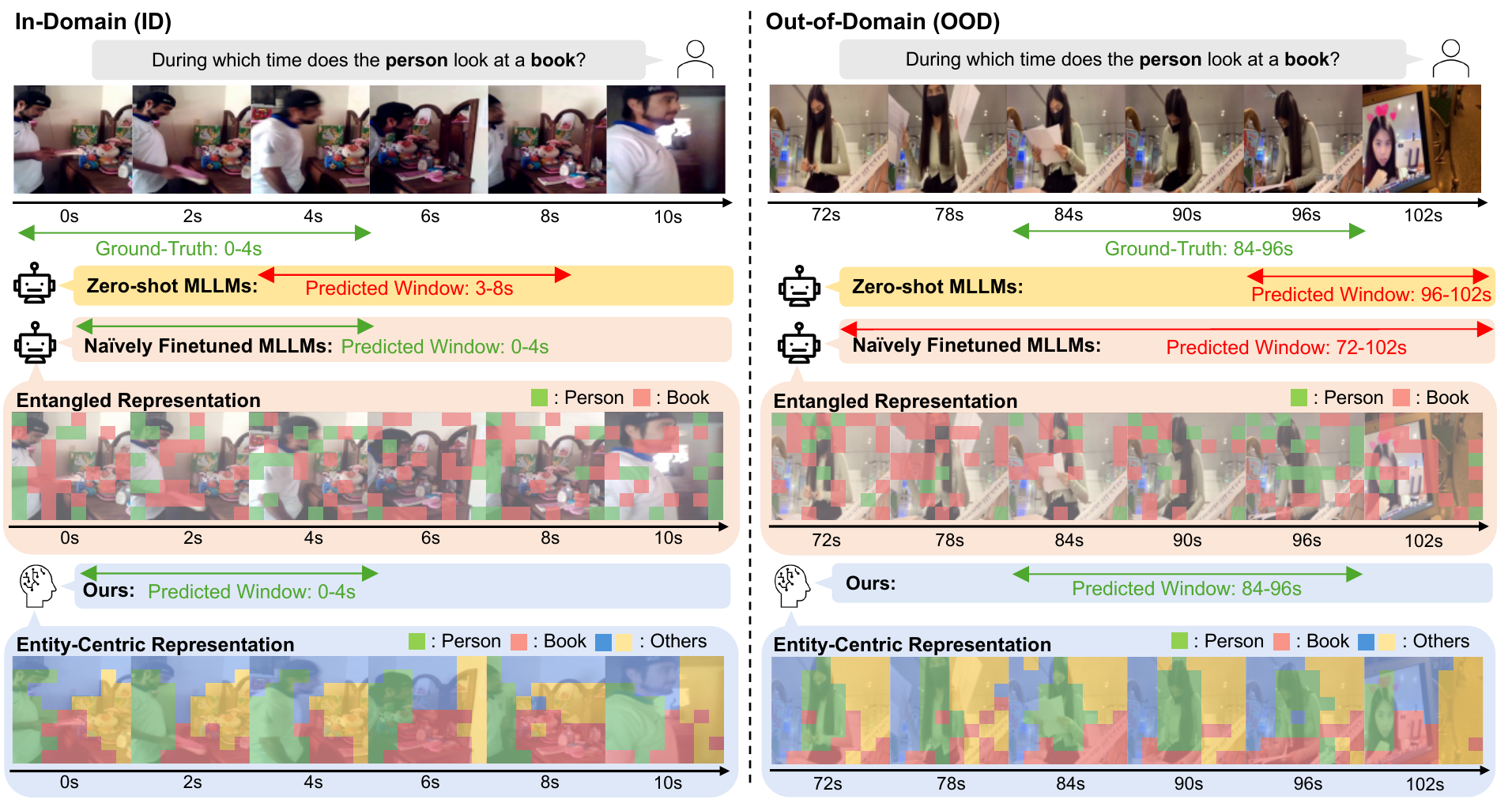}
\captionof{figure}{\textbf{Teaser.}
In this work, we tackle the domain generalization problem of MLLM-based VTG. 
We observe that na\"ively fine-tuned MLLMs learn entangled visual representations that rely on dataset-specific scene context rather than the actual visual content, indiscriminately responding to memorized patterns across frames in OOD videos.
In contrast, our method learns an \textit{entity-centric} visual representation that regularizes VTG fine-tuning to ground its predictions in the \textit{visual evidence} of the video, yielding transferability to OOD domains.
As a result, ours successfully captures both \emph{person} and \emph{book} consistently across frames and enables correct grounding regardless of the domain.
}

\label{fig:teaser}
\end{center}

%% file: 1_introduction.tex
\section{Introduction}
\label{sec:intro}

Video Temporal Grounding (VTG) requires localizing the temporal extent of an event described by a natural language query in an untrimmed video. 
While the task has been studied extensively with specialist temporal localization models such as DETR-based architectures~\cite{lei2021detecting,moon2023query,lin2023univtg,sun2024tr}, Multimodal Large Language Models (MLLMs) have recently emerged as a compelling alternative~\cite{huang2024vtimellm,ren2024timechat,guo2024trace,wang2024hawkeye,meinardus2024chrono,zeng2025timesuite}. 
The appeal of MLLMs is not merely their scale. 
Through large-scale image-text and video-text pre-training, they acquire broad visual-language alignment that can associate language queries with relevant visual entities across diverse video domains.

However, VTG exposes a critical \emph{adaptation paradox}. 
Before VTG-specific fine-tuning, pretrained MLLMs exhibit coarse but domain-agnostic \emph{entity-attention capability}: given a query, they can often attend to semantically relevant visual content across different visual distributions (\figref{observation} (a)).
Yet this zero-shot attention is too coarse for VTG, which demands exact start and end timestamps, making task-specific fine-tuning necessary~\cite{huang2024vtimellm,ren2024timechat,wang2024hawkeye,zeng2025timesuite,meinardus2024chrono}.
However, the extremely high cost of VTG annotations limits the scale and diversity of supervision, leaving source-domain fine-tuning vulnerable to shortcut learning—including temporal location bias~\cite{chae2024towards,otani2020uncovering,hao2022can}, query text bias~\cite{chae2024towards,li2022compositional,jiang2024mgqp}, and appearance bias~\cite{bao2022debiasttl,qi2024bssard}. 
These shortcuts cause fine-tuning to rely on source-specific patterns rather than the underlying entity-attention, so the learned localization knowledge fails to inherit the MLLM's domain-agnostic entity-attention.
As a result, fine-tuning improves In-Domain (ID) localization but harms generalization under Out-of-Domain (OOD) scenarios where shortcuts no longer hold (\figref{teaser}, \figref{observation} (b)).
We call this phenomenon 
\emph{attention-localization decoupling} 
in OOD VTG.

In this work, we find that this decoupling is primarily driven by \emph{visual domain shift}.
We decompose the OOD gap into two factors: a \emph{concept domain gap}, which captures whether the query involves subjects or actions unseen during source training, 
and a \emph{visual domain gap}, which measures how visually distant the OOD videos are from the source distribution. 
Splitting OOD samples by query-concept overlap yields only a marginal performance difference (\figref{diagnosis} (a)).
In contrast, splitting them by visual similarity to the source domain yields a substantially larger gap (\figref{diagnosis} (b)). 
These observations indicate that OOD degradation stems primarily from \emph{visual domain shift}.

These findings point to a different \emph{design principle} for cross-domain MLLM-based VTG: 
robust adaptation should anchor temporal localization in the \emph{inherent entity-attention} of the pre-trained MLLM, rather than merely fitting dense visual tokens to source-domain scene context.
To this end, we propose \textbf{\ours{}}, a parameter-efficient framework for \textbf{EVID}ence-grounded \textbf{ENT}ity-attentive adaptation for cross-domain VTG. 
Our goal is to learn transferable VTG knowledge by deriving temporal grounding predictions from \emph{query-relevant visual entity evidence}.
Realizing this principle requires three questions: (i) how to route adaptation through entities, (ii) what makes the entities meaningful, and (iii) how to leverage this entity information as evidence for temporal localization.

\ours{} answers these questions with three key components.
First, we propose an \textbf{Entity Bottleneck Adapter (\adapter{})}, which routes intermediate MLLM dense visual tokens through a compact set of entity slots to obtain domain-invariant representations.
Unlike prior work~\cite{slotvlm, chi2025slot, lei2026core} that re-runs instruction tuning to reshape MLLM representations into an entity-level form, our adapter design enables VTG learning grounded in entity-based visual representations at minimal cost. 
Second, we incorporate an \textbf{Entity-Binding Distillation (\dinoloss{})} loss to encourage each slot to bind to a \emph{coherent} entity.
We find that the pre-trained MLLM visual space alone has entangled representations and thus struggles to encode distinct entity information (\figref{teaser}, \figref{analysis}).
In contrast, \ours{} trained with our \dinoloss{} loss captures semantically coherent entities even under domain shift, as illustrated in \figref{teaser}. 
Third, to leverage the entity information captured by \adapter{} with \dinoloss{} as \emph{explicit visual evidence} guiding temporal localization predictions, we introduce an \textbf{Entity-to-eVidence (E2V) gating} mechanism.
Using the entity-attention of the pre-trained MLLM, we score each frame for the presence of query-relevant entities and modulate \adapter{}'s per-frame contribution accordingly. 
This injects an inductive bias that the target moment is where the query entities co-occur, enabling evidence-grounded localization under OOD shifts.

We evaluate \ours{} under cross-domain VTG settings, training on a single source dataset and evaluating on different target datasets~\cite{gao2017tall, lei2021detecting, anne2017localizing}. 
Across standard benchmarks, \ours{} consistently improves OOD robustness while maintaining competitive ID performance with modest parameter and memory overhead. 
These results support our central claim: for robust MLLM-based VTG under visual domain shift, what is needed is evidence-grounded adaptation that fully leverages the pre-trained entity-attention, not direct fitting to source-domain context.

We summarize our contributions as follows:
\begin{itemize}
    \item Through systematic analyses, we find that naïvely fine-tuning MLLMs causes \emph{attention-localization decoupling}—the learned temporal localization fails to inherit the pre-trained entity-attention capability—and show that it is primarily driven by \emph{visual domain shift}.
    \item We propose \ours{}, a parameter-efficient framework for evidence-grounded entity-attentive adaptation in cross-domain VTG. 
    Our entity bottleneck adapter with entity-binding distillation reshapes MLLM features into \textit{entity-centric} representations without re-running instruction tuning, while the proposed entity-to-evidence gating leverages these entities as \textit{explicit visual evidence for} temporal localization.
    \item We show that our evidence-grounded VTG adaptation yields consistent OOD robustness gains on diverse cross-domain VTG settings, while maintaining competitive in-domain performance with a modest number of parameters. 
\end{itemize}

%% file: 2_prior.tex
\section{Related Work}
\label{sec:related}

\paragraph{Video Temporal Grounding}
Video Temporal Grounding (VTG)~\cite{zhang2020hammer} aims to localize the temporal segments in an untrimmed video that correspond to a natural language query. 
Recent work~\cite{huang2024vtimellm, ren2024timechat} reformulates VTG as a generative task on MLLMs that produce temporal boundaries as text tokens via task-specific instruction tuning.
Follow-up work refines this formulation through interleaved frame–timestamp representations~\cite{meinardus2024chrono, li2025unitime}, causal event modeling~\cite{guo2024trace}, grounded tuning for long videos~\cite{zeng2025timesuite}, chain-of-LoRA reasoning~\cite{liu2026videomind}, and reinforcement learning with verifiable temporal rewards~\cite{wang2025timer}, all primarily relying on large-scale instruction tuning.
While these efforts enhance the temporal capability of MLLMs, the fine-tuning step that is inevitably required for accurate localization remains an unexamined source of risk: these methods do not directly address the OOD generalization failure that this fine-tuning induces, which is the main issue we study.

\paragraph{Object-centric Learning }
Object-centric learning represents a scene as a composition of entity-level parts, yielding transferable representations under domain shifts.
Slot attention~\cite{locatello2020object} is a representative mechanism, where learnable slots iteratively compete to explain the input, and has been extended to real-world images~\citep{seitzer2022bridging,lee2026divt} and videos~\citep{kipf2021conditional, wu2022slotformer, singh2022steve, wu2023slotdiffusion}. 
Recent work brings this idea into MLLMs: Slot-VLM~\cite{slotvlm} splits video tokens into object- and event-centric slots, Slot-MLLM~\cite{chi2025slot} combines a Q-Former with slot attention for unified multimodal generation, and CORE~\cite{lei2026core} compresses visual tokens into a compact set of slots for efficient MLLM inference, using segmentation masks from an off-the-shelf model as an object-centric prior.
However, all of these approaches require training the entire vision-language pipeline from scratch, whereas \ours{} introduces a parameter-efficient \adapter{} attachable to pre-trained MLLMs, bringing entity-centric structure while preserving the pretrained vision-language alignment.

%% file: 3_preliminaries.tex
\section{Problem Analysis}
\label{sec:background}

\input{figure/fig_observation}

To understand why na\"ively fine-tuned MLLMs suffer under domain shift, we conduct a series of empirical analyses.
We fine-tune Qwen2.5-VL-3B~\cite{bai2025qwen25vl} and InternVL3-2B~\cite{zhu2025internvl3} on Charades-STA (Cha.)~\cite{gao2017tall} and evaluate on QVHighlights (QVH)~\cite{lei2021detecting}.
See the appendix for the InternVL3~\cite{zhu2025internvl3} results and more details.

\subsection{OOD Failure Observations}
\label{sec:observations}

\noindent\textbf{Pre-trained MLLMs exhibit strong entity-attention capability.}
We first measure the visual-text alignment of the pre-trained Qwen2.5-VL-3B~\cite{bai2025qwen25vl} by computing the cosine similarity between visual patches and object text tokens. 
As shown in \figref{observation} (a), the pre-trained model exhibits strong visual-text alignment across visually distinct domains (Cha.~\cite{gao2017tall} and QVH~\cite{lei2021detecting}), with high similarity between object text tokens (\eg `book') and the corresponding visual patches at the frame level. 
This indicates that the inherent \textit{entity-attention} ability of pre-trained MLLMs to associate language queries with relevant visual content already transfers across diverse video distributions.

\noindent\textbf{Na\"ive fine-tuning causes severe OOD performance degradation.}
While the pre-trained model shows transferable alignment, na\"ively fine-tuning it on a single source dataset breaks this transferability (\figref{observation} (b)). 
Fine-tuning on QVH~\cite{lei2021detecting} produces a 19.8-point ID-OOD gap on Cha.~\cite{gao2017tall}, and fine-tuning on Cha.~\cite{gao2017tall} produces a 33.4-point gap on QVH~\cite{lei2021detecting}. 
The fine-tuned model overfits to source-domain patterns, failing to generalize to unseen domains.

\noindent\textbf{Acquired grounding capability fails to transfer.}
To probe the source of this collapse, we measure how strongly the fine-tuned model's visual-to-query cross-attention shifts toward ground-truth-interval frames, relative to the zero-shot baseline (\figref{observation} (c)). 
We focus on layers 10–18, following prior findings that visual-language interaction in VideoLLMs primarily occurs at intermediate layers~\cite{maptheflow}. 
Fine-tuning shifts attention toward ground-truth segments substantially on the trained domain (ID) but only marginally on the unseen domain (OOD), revealing that the learned localization does not transfer across domains.

\subsection{Domain Gap Diagnosis}
\label{sec:diagnosis}

We next ask: \emph{what is the primary cause of this transfer failure?} We decompose the domain gap into two sources, a \emph{concept domain gap} (whether OOD samples involve query concepts unseen during training) and a \emph{visual domain gap} (how visually distant OOD samples are from the source distribution), and analyze which one dominates.

\noindent\textbf{Concept gap does not explain OOD failure.}
We split the OOD test set (QVHighlights~\cite{lei2021detecting}) by query-concept overlap with ID (Charades-STA~\cite{gao2017tall}), separating samples whose query concepts appear in the source dataset from those that do not. 
As shown in \figref{diagnosis} (a), the performance gap between the two splits is marginal (3.3 points), indicating that pre-trained MLLMs already cover a broad range of concepts and the concept gap is not the dominant factor in OOD failure.

\noindent\textbf{Visual gap is the primary cause of OOD degradation.}
We instead split OOD samples by their visual feature similarity to the training set, where features are extracted from the vision encoder. 
As shown in \figref{diagnosis} (b), the most similar 20\% of OOD samples achieve 52.8 R1@0.5, while the most dissimilar 20\% drop to 39.1---a 13.7-point gap that scales with visual distance. 
This identifies the visual domain gap as the primary cause of OOD performance degradation.

\input{figure/fig_diagnosis}

\noindent\textbf{The model ignores visual content under domain shift.}
To verify whether the model genuinely grounds its predictions in visual content, we inject Gaussian noise into either the ground-truth (G.T.) interval or a random non-G.T. interval of the same length and measure R1@0.7 drop. 
As shown in \figref{diagnosis} (c), on ID, G.T. perturbation causes a much larger drop than random (17.4\% vs.\ 9.6\%), while on OOD the two are nearly identical (12.6\% vs.\ 12.1\%)—the model no longer attends to the actual visual content under distribution shift. 
Conversely, injecting noise into the entire non-G.T. interval improves performance on both ID and OOD (+13.1\% and +4.6\%).
It suggests that explicitly steering the model toward the visual content of the G.T. interval recouples its learned localization with the pre-trained entity-attention disrupted by visual domain shift, yielding gains even under OOD.

\vspace{-.3em}

%% file: figure/fig_observation.tex
\begin{figure}[t]
  \centering
  \includegraphics[width=\linewidth]{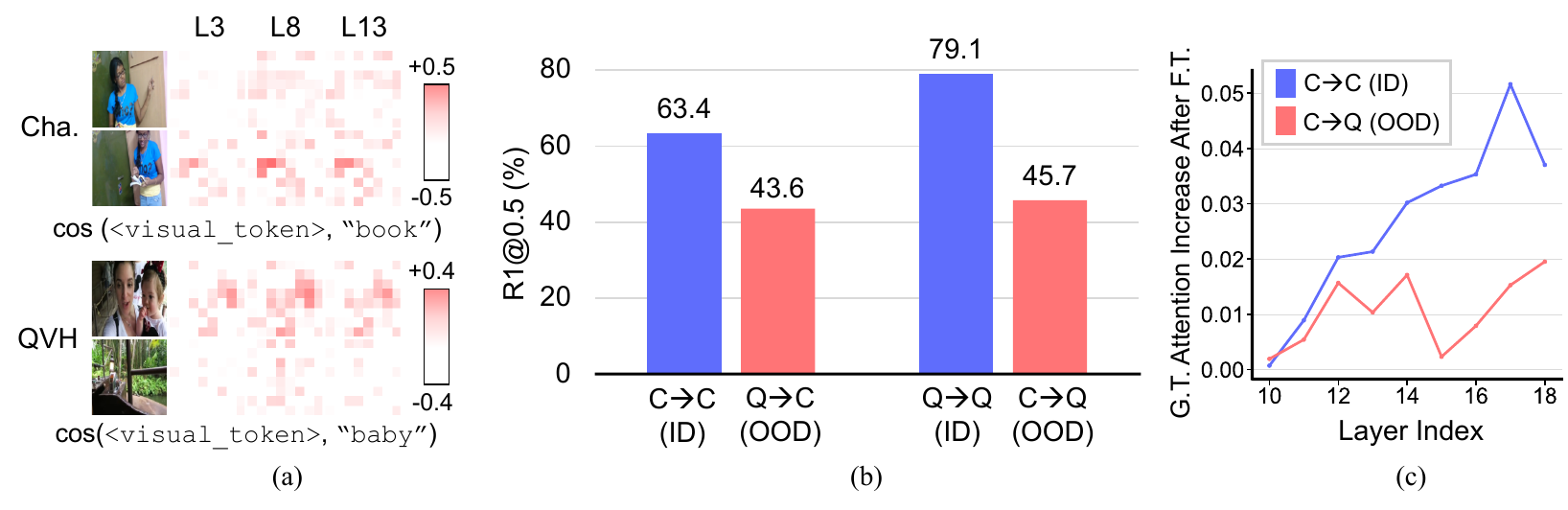}

\vspace{-.7em}
\caption{\textbf{Na\"ive fine-tuning breaks generalization.} 
(a) Cosine similarity between visual patches and object text tokens from Qwen2.5-VL-3B~\cite{bai2025qwen25vl} on Charades-STA (Cha.)~\cite{gao2017tall} and QVHighlights (QVH)~\cite{lei2021detecting}. Even zero-shot, the MLLM exhibits strong visual-text alignment regardless of video domain.
(b) When fine-tuned on a specific domain, the model suffers severe OOD drops.
(c) The model's attention on GT-interval frames relative to all frames increases substantially on the source domain (ID) but only marginally on the unseen domain (OOD), indicating that the learned VTG capability fails to transfer to other domains.
}
\vspace{-1em}
\label{fig:observation}
\end{figure}

%% file: figure/fig_diagnosis.tex
\begin{figure}[t]
\centering
\includegraphics[width=\linewidth]{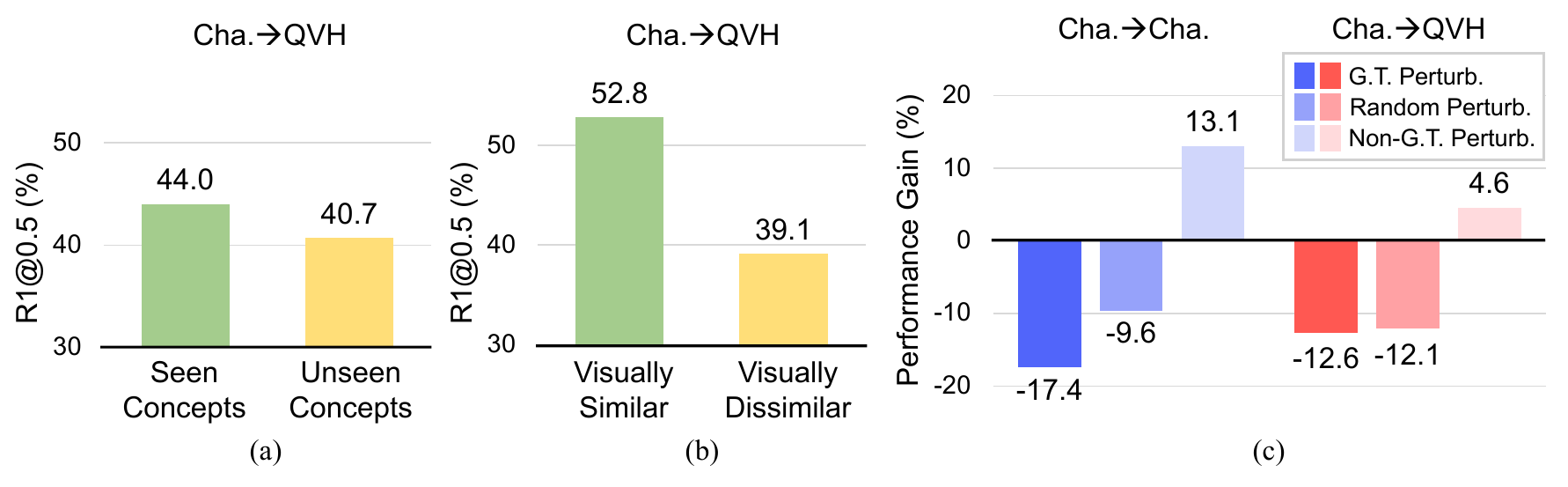}
\\
\vspace{-.7em}
\caption{\textbf{Visual domain gap dominates.} 
We decompose the domain gap into \textit{concept domain gap} and \textit{visual domain gap}.
(a) Splitting OOD samples by query-concept overlap with ID yields only a marginal gap.
(b) Ranking OOD samples by visual similarity to ID, the top-20\% most similar samples substantially outperform the bottom-20\%, identifying the \textit{visual domain gap} as the primary cause of degradation.
(c) On ID, perturbing the G.T. interval causes a much larger drop than perturbing a random non-G.T. interval of the same length, while on OOD the two yield nearly identical drops.
}
\vspace{-1em}
\label{fig:diagnosis}
\end{figure}

%% file: 4_method.tex
\section{Method}
\label{sec:method}

We introduce \textbf{\ours{}}, a parameter-efficient framework for cross-domain VTG that anchors temporal grounding in the inherent entity-attention of a pre-trained MLLM.
\ours{} builds entity-level structure in the early decoder layers, so that the later layers can apply this entity-attention without relying on domain-specific scene context.
To this end, the Entity Bottleneck Adapter (\textbf{\adapter{}}, \secref{slot_adapter}) routes dense visual tokens into entity slots, and the Entity-Binding Distillation (\textbf{\dinoloss{}}, \secref{sa_loss}) loss supplies the entity-level prior that the MLLM's visual space lacks.
Since slot attention operates only on visual tokens, the slots focus on separating visual content at the entity level, while query relevance is mainly handled by the pre-trained entity-attention through two paths:
Entity-to-eVidence gating (\textbf{E2V gating}, \secref{frame_gating}), which scores each frame by visual-text similarity to the query entities and modulates the \adapter{} output accordingly, and the vision-language integration in the middle and later layers~\citep{maptheflow}.
\figref{main} provides an overview.

\subsection{VTG Task Formulation}
\label{sec:framework}
\noindent\textbf{Interleaved Timestamps.} We follow the generative Video Temporal Grounding (VTG) paradigm, where an MLLM directly generates target timestamps as text tokens.
Given $T$ uniformly sampled video frames and a natural language query, we encode each frame into $N$ visual tokens $\mathbf{f}_t \in \mathbb{R}^{N \times D}$ via a frozen vision encoder and a linear projector, where $D$ is the hidden dimension of the LLM decoder, and tokenize its timestamp into a short text sequence $\mathbf{\tau{}}_t$.
The input $\mathbf{X}$ to the LLM decoder is constructed by interleaving each frame's visual tokens with its timestamp tokens, followed by the query tokens $\mathbf{q}$:
\begin{align}
    \mathbf{X} &= [\mathbf{f}_1, \mathbf{\tau{}}_1, \mathbf{f}_2, \mathbf{\tau{}}_2, \dots, \mathbf{f}_T, \mathbf{\tau{}}_T, \mathbf{q}],
    \label{eq:input_sequence}
\end{align}
This interleaved layout has been shown to be effective for temporal grounding~\cite{meinardus2024chrono,li2025unitime,zhang2025timelens}.
The model autoregressively decodes the target temporal window $[t_{\text{start}}, t_{\text{end}}]$.

\input{figure/fig_overview}

\subsection{Entity Bottleneck Adapter}
\label{sec:slot_adapter}
We propose Entity Bottleneck Adapter (\adapter{}), an adapter inserted into the MLLM decoder layers that reorganizes dense visual tokens into entity-based representations using a compact set of abstract slots.
By leaving the original vision-language inference path and instruction alignment intact, \adapter{} \emph{preserves the pre-trained entity-attention of the MLLM} while replacing only the domain-fragile dense visual representations with robust slot-based ones at minimal cost.

\noindent\textbf{Down Projection.}
We first project the entire input sequence $\mathbf{X}$ into a lower-dimensional bottleneck space: $\mathbf{X}_{\text{down}} = \mathbf{X}\mathbf{W}_{\text{down}} \in \mathbb{R}^{L \times d},$
where $\mathbf{W}_{\text{down}} \in \mathbb{R}^{D \times d}$ and $d \ll D$. 
From $\mathbf{X}_{\text{down}}$, we denote the visual portion as $\mathbf{X}_{\text{down}}^{\text{vis}} = [\mathbf{f}'_1, \dots, \mathbf{f}'_T] \in \mathbb{R}^{T \times N \times d}$ and the text portion as $\mathbf{X}_{\text{down}}^{\text{txt}} = [\mathbf{t}'_1, \dots, \mathbf{t}'_T, \mathbf{q}']$.

\noindent\textbf{Slot Attention Block.}
A set of $N_s$ learnable slot queries $\mathbf{S}^{(0)} \in \mathbb{R}^{N_s \times d}$ attend to the down-projected visual tokens $\mathbf{f}'_t \in \mathbb{R}^{N \times d}$ of each frame independently through $I$ iterations.
At iteration $i \in \{1, \dots, I\}$, we project the slots and the per-frame visual tokens into a common space: $\mathbf{Q}_t^{(i)} = \mathbf{S}_t^{(i-1)} \mathbf{W}_Q \in \mathbb{R}^{N_s \times d}$, $\mathbf{K}_t = \mathbf{f}'_t \mathbf{W}_K \in \mathbb{R}^{N \times d}$
, and 
$\mathbf{V}_t = \mathbf{f}'_t \mathbf{W}_V \in \mathbb{R}^{N \times d}$, where $\mathbf{W}_Q$, $\mathbf{W}_K$, and $\mathbf{W}_V$ are the projection matrices shared across frames. 
Next, the attention scores for frame $t$ at iteration $i$ are computed as: 
\begin{align}
\mathbf{M}_t^{(i)} &= \mathbf{K}_t (\mathbf{Q}_t^{(i)})^\top / \sqrt{d} \in \mathbb{R}^{N \times N_s}.
\end{align}
We normalize $\mathbf{M}_t^{(i)}$ along the \emph{slot axis} via softmax, fostering competitive assignment of visual tokens to slots within each frame:
\begin{align}
\mathbf{A}_t^{(i)}(n, k) &= \frac{\exp(\mathbf{M}_t^{(i)}(n, k))}{\sum_{j=1}^{N_s} \exp(\mathbf{M}_t^{(i)}(n, j))}.
\end{align}
We then normalize $\mathbf{A}_t^{(i)}$ along the token axis such that $\hat{\mathbf{A}}_t^{(i)}(\cdot, k)$ sums to one.
The aggregated update is computed as a weighted mean $\mathbf{Z}_t^{(i)} = (\hat{\mathbf{A}}_t^{(i)})^\top \mathbf{V}_t \in \mathbb{R}^{N_s \times d}$, and the slots are then updated via a Gated Recurrent Unit (GRU)~\cite{cho2014gru} based recurrence: $\mathbf{S}_t^{(i)} = \mathrm{GRU}(\mathbf{S}_t^{(i-1)}, \mathbf{Z}_t^{(i)})$.
This competition mechanism encourages each slot to specialize in a distinct semantic entity.

\noindent\textbf{Token Reconstruction.}
Since the LLM decoder expects the original token sequence length, we reconstruct the visual tokens by reusing the slot attention's assignment in a parameter-free manner. 
This design enforces that every reconstructed visual token is expressed only as a combination of slot features, thereby constraining the representations consumed by the LLM decoder to strictly adhere to \emph{disentangled entity-level semantics}.

Given the final slot representations $\mathbf{S}_t^{(I)} \in \mathbb{R}^{N_s \times d}$ for frame $t$, we reconstruct token-level features as a weighted sum of slot features based on the slot attention's assignment weights: 
\begin{align}
    \tilde{\mathbf{f'_t}} &= \hat{\mathbf{A}}_t^{(I)} \, \mathbf{S}_t^{(I)} \in \mathbb{R}^{N \times d}.
    \label{eq:attn_transpose}
\end{align}
We empirically find that using $\hat{\mathbf{A}}$ instead of $\mathbf{A}$ performs slightly better (see \secref{supp:ablation}).
We then concatenate the reconstructed visual portion $\tilde{\mathbf{X}}_{\text{down}}^{\text{vis}} = [\tilde{\mathbf{f}}'_1, \dots, \tilde{\mathbf{f}}'_T]$ with the unchanged text portion $\mathbf{X}_{\text{down}}^{\text{txt}}$ and project the result back to the original dimension via an up projection. The adapter output is added to the original input via a residual connection:
\begin{equation}
\mathbf{X}_{\text{out}} = \mathbf{X} + \mathtt{concat}(\tilde{\mathbf{X}}_{\text{down}}^{\text{vis}}, \mathbf{X}_{\text{down}}^{\text{txt}})\mathbf{W}_{\text{up}},
\end{equation}
where $\mathbf{W}_{\text{up}} \in \mathbb{R}^{d \times D}$ is initialized to zero, so the adapter acts as an identity mapping at the start of training.
This ensures training stability while gradually steering the representations toward entity-level decomposition~\cite{hu2022lora, yang2023aim}.

\noindent\textbf{Early-Layer Insertion.}
We attach the \adapter{} only to the early decoder layers.
Recent findings~\cite{maptheflow} show that cross-frame interactions occur in these early layers, while deeper layers handle language integration and answer generation.
By inserting the \adapter{} at this stage, each slot captures temporally coherent semantics across frames rather than frame-independent decompositions.
The deeper layers, fine-tuned with LoRA~\cite{hu2022lora,yang2023aim}, then reason over these disentangled representations.

\subsection{Entity-Binding Distillation}
\label{sec:sa_loss}

While the \adapter{} encourages decomposition through its bottleneck structure, the bottleneck alone is insufficient: the entangled visual space of the \naive{} MLLM lacks the entity-level structure for slots to bind to.
As a result, slots fail to specialize and produce near-uniform attention weights over visual tokens, as shown in \secref{additional_results}.
To address this, we employ an Entity-Binding Distillation (\dinoloss{}) loss.
Inspired by prior works on object-centric learning~\cite{seitzer2022bridging, akan2025slotdiff}, our loss distills the objectness prior from a self-supervised vision model to encourage semantically coherent slot formation, as illustrated in \figref{main} (b).

\noindent\textbf{Entity-level Matching.}
We extract DINOv2~\citep{oquab2023dinov2} patch features from $T$ frames, apply $2\times 2$ average pooling to match the MLLM's spatial resolution, and $\ell_2$-normalize along the feature dimension. 
We then perform $K$-means clustering ($K=N_s$) on each frame independently, 
yielding a binary cluster map $\mathbf{C} \in \{0,1\}^{T \times N_s \times N}$, where $\mathbf{C}_{t,k,n}=1$ indicates that patch $n$ in frame $t$ belongs to cluster $k$.
Since slot and cluster indices are arbitrary, we perform per-frame Hungarian matching~\citep{kuhn1955hungarian} between the normalized slot attention maps $\hat{\mathbf{A}} \in [0,1]^{T \times N_s \times N}$ and $\mathbf{C}$, using patch-averaged BCE as the cost, to obtain an optimal one-to-one matching $\pi^{(t)}: \{1,\dots,N_s\} \rightarrow \{1,\dots,N_s\}$.

\noindent\textbf{Loss.}
Given the matching, we define the alignment loss as the BCE between matched slot--cluster pairs, averaged over all frames, slots, and patches:
\begin{equation}
L_{\text{EBD}} = -\frac{1}{TN_sN} \sum_{t=1}^{T} \sum_{k=1}^{N_s} \sum_{n=1}^{N}
\left[ \mathbf{C}^{(t)}_{\pi^{(t)}(k), n} \log \hat{\mathbf{A}}^{(t)}_{k, n} +
\left(1 - \mathbf{C}^{(t)}_{\pi^{(t)}(k), n}\right) \log \left(1 - \hat{\mathbf{A}}^{(t)}_{k, n}\right) \right]. \\
\label{eq:EBD_loss}
\vspace{-1em}
\end{equation}

\subsection{Entity to Evidence Gating}
\label{sec:frame_gating}
We introduce an Entity-to-eVidence (E2V) gating mechanism to leverage the entities captured by \adapter{} as explicit evidence for temporal grounding, injecting a VTG inductive bias that \textit{temporal segments where query-relevant entities co-occur are more likely to be the target moment}. Specifically, we use the MLLM's pretrained entity-attention to score per-frame entity presence and modulate \adapter{}'s output accordingly, as shown in \figref{main} (c).

\noindent\textbf{Entity Extraction from Query.}
For each natural language query, we extract the subject and object concept words as query-relevant entities.
We perform this offline using a pre-trained LLM (Qwen3-4B~\cite{yang2025qwen3}) as a preprocessing step.
We denote the extracted indices in the query token sequence as $i_{\text{sub}}$ and $i_{\text{obj}}$. See \secref{supp_concept_extraction} for more details.
Using these indices, we select the corresponding query tokens 
$\mathbf{q'}_{\text{sub}}, \mathbf{q'}_{\text{obj}} \in \mathbb{R}^d$ from the down-projected text portion $\mathbf{X}_{\text{down}}^{\text{txt}}$.

\noindent\textbf{Frame-level Co-occurrence Score.}
We compute the cosine similarity between each frame's averaged visual token and the subject/object query tokens: 
$s^{\text{sub}}_t = \cos(\bar{\mathbf{f}}_t, \mathbf{q'}_{\text{sub}}), \quad s^{\text{obj}}_t = \cos(\bar{\mathbf{f}}_t, \mathbf{q'}_{\text{obj}}),$
where $\bar{\mathbf{f}}_t \in \mathbb{R}^d$ is a per-frame visual representation obtained by averaging the $N$ down-projected visual tokens $\mathbf{X}_{\text{down}}^{\text{vis}}$.
We then apply min-max normalization, yielding $\hat{s}^{\text{sub}}_t$ and $\hat{s}^{\text{obj}}_t$.
The final gating score for frame $t$ is the product of the two normalized scores: 
\begin{equation}
g_t = \hat{s}^{\text{sub}}_t \cdot \hat{s}^{\text{obj}}_t \in [0, 1].
\end{equation}
This product captures the co-occurrence of the subject and object: $g_t$ is high only when both entities are present in frame $t$.

\noindent\textbf{Gated Reconstruction.}
We modulate the reconstructed visual tokens by broadcasting the per-frame gating scores $\mathbf{g} = [g_1, \dots, g_T]$ across the $N$ tokens within each frame, and replace the up projection step in the \adapter{} accordingly: $\mathbf{X}_{\text{out}} = \mathbf{X} + \mathtt{concat}(\mathbf{X}_{\text{down}}^{\text{vis}} \odot \mathbf{g}, \mathbf{X}_{\text{down}}^{\text{txt}})\mathbf{W}_{\text{up}},$
where $\odot$ denotes element-wise multiplication with broadcasting along the token and feature dimensions. 
As a result, the \adapter{}'s contribution is concentrated on frames where query-relevant entities co-occur.

\subsection{Training Objective}
\label{sec:loss}
The framework is trained end-to-end with the vision encoder frozen. The \adapter{} and LoRA~\cite{hu2022lora} parameters are updated jointly. 
We train \ours{} with the combined objective $L_{\text{total}} = L_{\text{CE}} + \lambda L_{\text{EBD}},$
where $\lambda$ controls the strength of the objectness prior.

%% file: figure/fig_overview.tex
\begin{figure}[t]
\centering
\includegraphics[width=\textwidth]{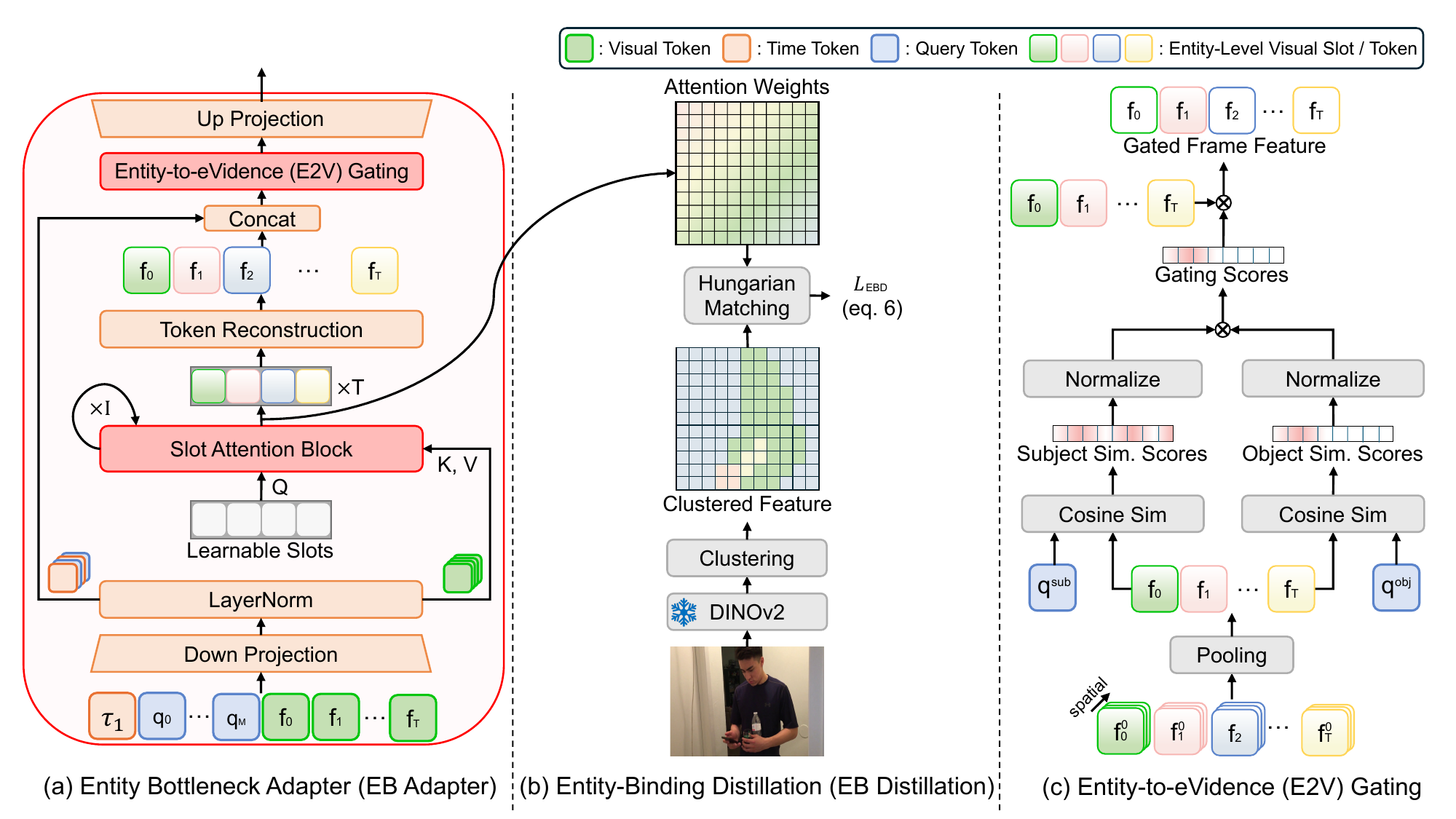}

\caption{
\textbf{Overview. (a) Entity Bottleneck Adapter (EB Adapter).} A lightweight adapter in the early LLM decoder layers decomposes visual tokens into entity-level slots via slot attention.
\textbf{(b) Entity-Binding Distillation (EB Distillation).} Entity cluster maps derived from DINOv2~\cite{oquab2023dinov2} patch features are distilled into slot assignment maps via BCE loss, encouraging semantically coherent slot formation.
\textbf{(c) Entity-to-eVidence (E2V) Gating.} A per-frame gating score, reflecting the co-occurrence of query-relevant entities, modulates the adapter's influence.
}
\vspace{-2em}
\label{fig:main}
\end{figure}

%% file: 5_results.tex
\input{table/main_all}

\section{Experiments}
\label{sec:results}

\subsection{Experimental Setup}

\noindent\textbf{Training \& Evaluation.}
We build on Qwen2.5-VL-Instruct~\cite{bai2025qwen25vl} (7B) as the backbone MLLM and fine-tune it on two source datasets, Charades-STA (Cha.)~\cite{gao2017tall} and QVHighlights (QVH)~\cite{lei2021detecting}.
We evaluate the models on three target datasets: Cha., QVH, and DiDeMo~\cite{anne2017localizing}. 
Thanks to the parameter-efficient design of \ours{}, only the \adapter{} and LoRA modules are trainable, amounting to approximately 201M parameters (\textit{2.42\% of the 7B backbone}). We denote each setting by its source-target pair (\emph{e.g.}, Cha.$\rightarrow$ QVH), reporting In-Domain (ID) performance when source equals target and Out-of-Domain (OOD) performance otherwise. All results are reported using the standard moment retrieval metrics R1@0.3, R1@0.5, R1@0.7, and mIoU.

\noindent\textbf{Baselines.}
We compare \ours{} against four categories of methods.
(1)~{Proprietary MLLMs}: GPT-5~\citep{OpenAI2025_GPT5} and Gemini-2.0-Flash~\citep{comanici2025gemini-2-5}, whose results are taken from Zhang et al.~\citep{zhang2025timelens}.
(2)~{Large-scale instruction-tuned MLLMs for VTG}, evaluated in a zero-shot manner: HawkEye~\cite{wang2024hawkeye}, TimeSuite~\cite{zeng2025timesuite}, UniTime~\cite{li2025unitime}, and VideoMind~\cite{liu2026videomind}.
(3)~{DETR-based specialists} trained on a single source dataset: EaTR~\cite{jang2023knowing} and CG-DETR~\cite{moon2023correlation}.
(4)~{MLLM-based methods} fine-tuned on a single source dataset: Chrono~\cite{meinardus2024chrono} with a BLIP-2~\cite{li2023blip2} backbone, and Qwen2.5-VL~\cite{bai2025qwen25vl} baseline fine-tuned with LoRA~\cite{hu2022lora} under the same training setup as ours.

\subsection{Cross-Domain Temporal Grounding Results}

\input{table/ablation}

\noindent\textbf{Comparison with Existing Methods.}
In \tabref{main}, \ours{} consistently improves OOD performance across all source-target configurations while maintaining competitive ID performance.
Among MLLM-based methods, we observe two failure modes that support our motivation presented in \secref{intro}.
First, zero-shot MLLMs, even those that are large-scale instruction-tuned, remain substantially below fine-tuned methods in the ID setting, confirming that task-specific adaptation is necessary for precise temporal localization.
Second, na\"ive source-domain fine-tuning suffers severely under domain shift: although Qwen2.5-VL~\cite{bai2025qwen25vl} reaches strong ID performance, its OOD R1@0.5 drops by over 19.6 points on Cha.$\rightarrow$QVH and over 23.9 points on QVH$\rightarrow$Cha.
\ours{} bridges these two regimes, delivering consistent OOD gains over the Qwen2.5-VL baseline (\emph{e.g.}, +5.5 R1@0.5 and +7.2 R1@0.7 on Cha.$\rightarrow$QVH).
\input{table/other_backbones}These results demonstrate that entity-grounded visual evidence provides a robust pathway to fully leverage the entity-attention capabilities of pretrained MLLMs under domain shift.

\noindent\textbf{Results with Other MLLM Backbones.}
In \tabref{other_backbones}, we further apply \ours{} to Qwen3-VL-8B~\cite{Qwen3-VL} and InternVL3.5-8B~\cite{wang2025internvl3_5}.
\ours{} improves OOD performance on both backbones while maintaining ID performance, with especially large gains on InternVL3.5-8B (+8.0 and +10.1 R1@0.5 on QVH and DiDeMo).
This shows that \ours{} is not tied to a specific MLLM architecture.

\subsection{Ablation Study}
\label{sec:ablation}

\noindent\textbf{Effects of \ours{} Components.}
We investigate the effect of each component of \ours{} in \tabref{ablation_ours} (a).
Comparing the results with and without \dinoloss{}, we observe that applying the \adapter{} alone to the semantically unstructured MLLM visual space fails to yield effective entity-centric representations, as discussed in \secref{sa_loss}.
Only when our \dinoloss{} strategy is incorporated does the model learn transferable entity-level visual representations, leading to a gain in R1@0.5 by 3.9 points under the OOD setting. 
On top of this, the VTG-oriented inductive bias injected by \gating{} provides an additional 1.0-point improvement in OOD R1@0.5, confirming that the two components are complementary.

\noindent\textbf{Effects of Adapter Design.}
In \tabref{ablation_ours} (b), we show that attaching the learnable module to the MLLM decoder in an adapter style is the most effective placement for entity-level restructuring while preserving the pre-trained entity-attention. 
Compared to a variant that inserts a slot attention block between the vision encoder and the LLM decoder, which suffers a catastrophic performance collapse, our adapter-style design avoids disrupting the MLLM's pre-trained alignment and still enables effective entity-level decomposition.

\noindent\textbf{Effects of Entity-level Decomposition.}
In \tabref{ablation_ours} (c), an adapter with standard self-attention improves ID R1@0.7 over LoRA~\citep{hu2022lora} but suffers a severe OOD drop (-11.9 R1@0.5), indicating overfitting. In contrast, our \adapter{} maintains the ID gain and improves OOD by +3.4 R1@0.7, demonstrating that entity-level decomposition learns domain-invariant visual representations.

\noindent\textbf{Effects of Query-dependent \gating{}.}
In \tabref{ablation_ours} (d), we compare two strategies for selecting the entities used to compute the gating score in \gating{}: extracting per-sample subject/object words from each query (query-dependent) versus using the fixed generic words, \emph{e.g.}, `person' and `object', for all samples (query-agnostic). 
The query-dependent variant yields a gain in OOD R1@0.5 by 1.4 points, demonstrating that our gating actively exploits the MLLM's inherent entity-attention.

\input{figure/fig_additional_results}
\subsection{Additional Analysis on \ours{}}
\label{sec:additional_results}

\noindent\textbf{How Does \dinoloss{} Affect Entity-Binding?}
In \figref{analysis} (a), we measure the entropy of each slot's attention over visual tokens and the pairwise cosine similarity between slots across all \adapter{} layers.
When \adapter{} is trained directly on the pre-trained MLLM visual space, the attention is nearly uniform (entropy $\approx 4.2$, the uniform bound) with near-identical slots (cosine similarity $\approx 1.0$), indicating that this space does not afford entity-level binding on its own.
With the \dinoloss{} loss, however, both metrics drop substantially (entropy $\approx 3.5$, cosine similarity $\approx 0.3$), showing that each slot binds to a distinct, localized entity.

\noindent\textbf{What Do Slots Learn?}
In \figref{analysis} (b), we visualize the slot attention maps of \ours{}.
Across both ID and OOD samples, the slots decompose scenes into semantically coherent regions such as people, objects, and backgrounds.
Importantly, this decomposition generalizes to unseen domains without any domain-specific supervision, confirming that \ours{} learns \textit{transferable entity-level representations} rather than source-biased patterns.

\noindent\textbf{Does \ours{} Resolve Attention-Localization Decoupling?}
In \figref{analysis} (c), we measure \ours{}'s attention increase on G.T. frames over to all frames, following the analysis in \figref{observation} (c).
In the later layers, \ours{}'s OOD increase becomes comparable to its ID increase, showing that \ours{} performs localization grounded in entity-attention even on unseen domains.

%% file: table/main_all.tex
\begin{table*}[t]
\centering
\caption{\textbf{Performance comparison on cross-domain video temporal grounding benchmarks.} 
We evaluate \ours{} on Charades-STA~\cite{gao2017tall}, QVHighlights~\cite{lei2021detecting}, and DiDeMo~\cite{anne2017localizing}, reporting both \textcolor{gray}{In-Domain (ID)} (source=target) and Out-of-Distribution (OOD) (source$\neq$target) settings.
$\dagger$ denotes our implemented baseline fine-tuned with LoRA~\cite{hu2022lora}. 
\colorbox{green!10}{Our results} are highlighted; \textbf{bold} marks the best under each source$\rightarrow$target setting. 
\colorbox{bggray}{Zero-shot} models are reported for reference, with cells left blank (-) for target datasets seen during their training.
}

\renewcommand{\arraystretch}{1.3}
\setlength{\tabcolsep}{3pt}
\resizebox{\textwidth}{!}{
\begin{tabular}{c l c cccc c cccc c cccc}
\toprule
\multirow{3}{*}{\shortstack{Source\\dataset}} & \multirow{3}{*}{Method} & \multirow{3}{*}{\shortstack{LLM\\size}} & \multicolumn{14}{c}{Target dataset} \\
\cmidrule{4-17}
& & & \multicolumn{4}{c}{Charades-STA} && \multicolumn{4}{c}{QVHighlights} && \multicolumn{4}{c}{DiDeMo} \\
\cmidrule{4-7} \cmidrule{9-12} \cmidrule{14-17}
& & & R1@0.3 & R1@0.5 & R1@0.7 & mIoU && R1@0.3 & R1@0.5 & R1@0.7 & mIoU && R1@0.3 & R1@0.5 & R1@0.7 & mIoU \\
\midrule
\multirow{6}{*}{\rotatebox[origin=c]{90}{Zero-Shot}} 
& \cellcolor{bggray}GPT-5~\cite{OpenAI2025_GPT5} & \cellcolor{bggray}- & \cellcolor{bggray}39.7 & \cellcolor{bggray}18.3 & \cellcolor{bggray}6.2 & \cellcolor{bggray}28.4 &\cellcolor{bggray}& \cellcolor{bggray}- & \cellcolor{bggray}- & \cellcolor{bggray}- & \cellcolor{bggray}- &\cellcolor{bggray}& \cellcolor{bggray}- & \cellcolor{bggray}- & \cellcolor{bggray}- & \cellcolor{bggray}- \\
& \cellcolor{bggray}{\fontsize{10pt}{11pt}\selectfont Gemini-2.0-Flash}~\cite{comanici2025gemini-2-5} & \cellcolor{bggray}- & \cellcolor{bggray}55.6 & \cellcolor{bggray}29.0 & \cellcolor{bggray}9.5 & \cellcolor{bggray}35.1 &\cellcolor{bggray}& \cellcolor{bggray}72.1 & \cellcolor{bggray}58.1 & \cellcolor{bggray}41.9 & \cellcolor{bggray}54.9 &\cellcolor{bggray}& \cellcolor{bggray}- & \cellcolor{bggray}- & \cellcolor{bggray}- & \cellcolor{bggray}- \\
& \cellcolor{bggray}HawkEye~\cite{wang2024hawkeye} & \cellcolor{bggray}7B & \cellcolor{bggray}50.6 & \cellcolor{bggray}31.4 & \cellcolor{bggray}14.5 & \cellcolor{bggray}33.7 &\cellcolor{bggray}& \cellcolor{bggray}42.1 & \cellcolor{bggray}20.8 & \cellcolor{bggray}6.6 & \cellcolor{bggray}28.7 &\cellcolor{bggray}& \cellcolor{bggray}37.8 & \cellcolor{bggray}18.4 & \cellcolor{bggray}7.8 & \cellcolor{bggray}24.7 \\
& \cellcolor{bggray}TimeSuite~\cite{zeng2025timesuite} & \cellcolor{bggray}7B & \cellcolor{bggray}69.9 & \cellcolor{bggray}48.7 & \cellcolor{bggray}24.0 & \cellcolor{bggray}- &\cellcolor{bggray}& \cellcolor{bggray}-  & \cellcolor{bggray}12.3 & \cellcolor{bggray}9.2 & \cellcolor{bggray}21.3 &\cellcolor{bggray}& \cellcolor{bggray}- & \cellcolor{bggray}- & \cellcolor{bggray}- & \cellcolor{bggray}- \\
& \cellcolor{bggray}UniTime~\cite{li2025unitime} & \cellcolor{bggray}7B & \cellcolor{bggray}- & \cellcolor{bggray}{59.1} & \cellcolor{bggray}{31.9} & \cellcolor{bggray}{52.2} &\cellcolor{bggray}& \cellcolor{bggray}- & \cellcolor{bggray}41.0 & \cellcolor{bggray}31.5 & \cellcolor{bggray}43.7 &\cellcolor{bggray}& \cellcolor{bggray}- & \cellcolor{bggray}- & \cellcolor{bggray}- & \cellcolor{bggray}- \\
& \cellcolor{bggray}VideoMind~\cite{liu2026videomind} & \cellcolor{bggray}7B & \cellcolor{bggray}73.5 & \cellcolor{bggray}{59.1} & \cellcolor{bggray}31.2 & \cellcolor{bggray}50.2 &\cellcolor{bggray}& \cellcolor{bggray}- & \cellcolor{bggray}- & \cellcolor{bggray}- & \cellcolor{bggray}- &\cellcolor{bggray}& \cellcolor{bggray}- & \cellcolor{bggray}- & \cellcolor{bggray}- & \cellcolor{bggray}- \\
\midrule
\multirow{5}{*}{\rotatebox[origin=c]{90}{Charades-STA}} 
& EaTR~\cite{jang2023knowing} & - & \textcolor{gray}{67.7} & \textcolor{gray}{55.2} & \textcolor{gray}{33.1} & \textcolor{gray}{47.7} && 31.7 & 17.0 & 6.4 & 21.5 && 32.7 & 20.1 & 7.8 & 21.2 \\
& CG-DETR~\cite{moon2023correlation} & - & \textcolor{gray}{69.7} & \textcolor{gray}{57.6} & \textcolor{gray}{35.1} & \textcolor{gray}{49.5} && 37.4 & 22.8 & 10.5 & 25.2 && 30.3 & 21.8 & 11.0 & 20.7 \\
& Chrono-BLIP~\cite{meinardus2024chrono}  & 4B & \textcolor{gray}{79.1} & \textcolor{gray}{68.9} & \textcolor{gray}{48.7} & \textcolor{gray}{58.6} && 67.2 & 44.5 & 23.7 & 44.3 && 56.0 & 43.2 & 23.1 & 37.7 \\
& Qwen2.5-VL$^\dagger$~\cite{bai2025qwen25vl} & 7B & \textcolor{gray}{80.3} & \textcolor{gray}{67.7} & \textcolor{gray}{43.9} & \textcolor{gray}{58.1} && 75.2 & 54.3 & 29.7 & 51.2 && 68.2 & 53.8 & 30.3 & 47.4 \\
& \cellcolor{green!10}\ours & \cellcolor{green!10}7B & \cellcolor{green!10}\textcolor{gray}{80.3} & \cellcolor{green!10}\textcolor{gray}{69.1} & \cellcolor{green!10}\textcolor{gray}{49.2} & \cellcolor{green!10}\textcolor{gray}{59.9} &\cellcolor{green!10}& \cellcolor{green!10}\textbf{77.5} & \cellcolor{green!10}\textbf{59.8} & \cellcolor{green!10}\textbf{36.9} & \cellcolor{green!10}\textbf{55.2} &\cellcolor{green!10}& \cellcolor{green!10}\textbf{69.2} & \cellcolor{green!10}\textbf{55.4} & \cellcolor{green!10}\textbf{32.3} & \cellcolor{green!10}\textbf{48.5} \\
\midrule
\multirow{5}{*}{\rotatebox[origin=c]{90}{QVHighlights}} 
& EaTR~\cite{jang2023knowing} & - & 40.8 & 27.2 & 13.0 & 28.0 && \textcolor{gray}{70.3} & \textcolor{gray}{59.6} & \textcolor{gray}{40.3} & \textcolor{gray}{53.1} && 36.0 & 17.8 & 6.5 & 24.1 \\
& CG-DETR~\cite{moon2023correlation} & - & 42.8 & 25.5 & 12.2 & 28.5 && \textcolor{gray}{77.5} & \textcolor{gray}{65.6} & \textcolor{gray}{52.1} & \textcolor{gray}{61.3} && 36.6 & 21.2 & 8.9 & 24.6 \\
& Chrono-BLIP~\cite{meinardus2024chrono}  & 4B & 63.3 & 39.3 & 20.4 & 42.7 && \textcolor{gray}{86.1} & \textcolor{gray}{76.6} & \textcolor{gray}{62.7} & \textcolor{gray}{70.7} && 43.0 & 24.4 & 11.8 & 33.7 \\
& Qwen2.5-VL$^\dagger$~\cite{bai2025qwen25vl} & 7B & 71.1 & 48.1 & 26.2 & 47.3 && \textcolor{gray}{86.9} & \textcolor{gray}{78.2} & \textcolor{gray}{64.8} & \textcolor{gray}{72.0} && 59.4 & 34.1 & 16.9 & 38.2 \\
& \cellcolor{green!10}\ours & \cellcolor{green!10}7B & \cellcolor{green!10}\textbf{73.3} & \cellcolor{green!10}\textbf{50.2} & \cellcolor{green!10}\textbf{28.0} & \cellcolor{green!10}\textbf{49.0} &\cellcolor{green!10}& \cellcolor{green!10}\textcolor{gray}{89.5} & \cellcolor{green!10}\textcolor{gray}{82.2} & \cellcolor{green!10}\textcolor{gray}{69.1} & \cellcolor{green!10}\textcolor{gray}{75.3} &\cellcolor{green!10}& \cellcolor{green!10}\textbf{62.4} & \cellcolor{green!10}\textbf{36.0} & \cellcolor{green!10}\textbf{17.4} & \cellcolor{green!10}\textbf{40.6} \\
\bottomrule
\end{tabular}
\label{tab:main}
    }
\vspace{-1em}
\end{table*}

%% file: table/ablation.tex
\begin{table*}[t]
\centering
\caption{\textbf{Ablation study.} 
To validate the effect of each component in \ours{}, we show the results on the Cha.$\rightarrow$QVH setting.
The best numbers are \textbf{highlighted}.
}

\mpage{0.49}{
{\fontsize{9pt}{11pt}\selectfont (a) Effects of \ours{} components.}
\\
\resizebox{\linewidth}{!}{
\begin{tabular}{ccc c cc c cc}
\toprule
\multirow{2}{*}{\adapter{}} & \multirow{2}{*}{\dinoloss{}} & \multirow{2}{*}{\gating{}} && \multicolumn{2}{c}{Cha. (ID)} && \multicolumn{2}{c}{QVH (OOD)} \\
\cline{5-6} \cline{8-9}
&&&& R1@0.5 & R1@0.7 && R1@0.5 & R1@0.7 \\
\midrule
& & && 67.7 & 43.9 && 54.3 & 29.7 \\ 
\checkmark & & && 68.1 & 49.5 && 54.9 & 33.1 \\ 
\checkmark & \checkmark & && \textbf{69.1} & \textbf{50.4} && 58.8 &  36.7 \\ 
\checkmark & \checkmark & \checkmark && \textbf{69.1} & 49.2 && \textbf{59.8} & \textbf{36.9} \\ 
\bottomrule
\end{tabular}
}
}
\mpage{0.49}{ 
{\fontsize{9pt}{11pt}\selectfont (b) Effects of adapter design.}
\\
\resizebox{\linewidth}{!}{
\begin{tabular}{l c cc c cc}
\toprule
Slot Attention && \multicolumn{2}{c}{Cha. (ID)} & 
& \multicolumn{2}{c}{QVH (OOD)} \\ 
\cline{3-4} \cline{6-7} 
Placement && R1@0.5 & R1@0.7 && R1@0.5 & R1@0.7 \\
\midrule
After vision encoder && 34.6 & 18.8 && 19.5 & 12.7 \\
Adapter-style (ours) && \textbf{68.1} & \textbf{49.5} && \textbf{54.9} & \textbf{33.1} \\ 
\bottomrule
\end{tabular}
}
}

\mpage{0.49}{
{\fontsize{9pt}{11pt}\selectfont (c) Effects of entity-level decomposition.}
\\
\resizebox{\linewidth}{!}{
\begin{tabular}{l c cc c cc}
\toprule
\multirow{2}{*}{Bottleneck Design} && \multicolumn{2}{c}{Cha. (ID)} & 
& \multicolumn{2}{c}{QVH (OOD)} \\ 
\cline{3-4} \cline{6-7} 
&& R1@0.5 & R1@0.7 && R1@0.5 & R1@0.7 \\
\midrule
LoRA && 67.7 & 43.9 && 54.3 & 29.7 \\
Adapter w/ self-attn. && \textbf{68.8} & \textbf{49.5} && 42.4 & 21.0 \\ 
\adapter{} && 68.1 & \textbf{49.5} && \textbf{54.9} & \textbf{33.1} \\ 
\bottomrule
\end{tabular}
}
}
\mpage{0.49}{
{\fontsize{9pt}{11pt}\selectfont (d) Effects of query-dependent \gating{}.}
\\
\resizebox{\linewidth}{!}{
\begin{tabular}{l c cc c cc}
\toprule
\multirow{2}{*}{Entity Source} && \multicolumn{2}{c}{Cha. (ID)} & 
& \multicolumn{2}{c}{QVH (OOD)} \\ 
\cline{3-4} \cline{6-7} 
&& R1@0.5 & R1@0.7 && R1@0.5 & R1@0.7 \\
\midrule
Query-agnostic && \textbf{69.3} & \textbf{49.4} && 58.4 & 36.0 \\
Query-dependent && 69.1 & 49.2 && \textbf{59.8} & \textbf{36.9} \\ 
\bottomrule
\end{tabular}
}
}

\vspace{-1.5em}
\label{tab:ablation}
\label{tab:ablation_ours}
\end{table*}

%% file: table/other_backbones.tex
\begin{wraptable}{r}{0.55\textwidth}
\centering
\vspace{-.8em}
\caption{\textbf{Performance with other backbones.} $\dagger$ denotes the fine-tuned baseline. \textbf{Bold} marks the best OOD result under each backbone.}
\vspace{-.5em}
\setlength{\tabcolsep}{2pt}
\resizebox{\linewidth}{!}{
\begin{tabular}{l cc cc cc}
\toprule
\multirow{2}{*}{Backbone}
& \multicolumn{2}{c}{Cha. (ID)}
& \multicolumn{2}{c}{QVH (OOD)}
& \multicolumn{2}{c}{DiDeMo (OOD)} \\
\cmidrule(lr){2-3} \cmidrule(lr){4-5} \cmidrule(lr){6-7}
& R1@0.5 & R1@0.7 & R1@0.5 & R1@0.7 & R1@0.5 & R1@0.7 \\
\midrule
Qwen3-VL-8B$^\dagger$ & \textcolor{gray}{70.4} & \textcolor{gray}{50.1} & 61.4 & 44.2 & 58.6 & \textbf{30.8} \\
\rowcolor{green!10}
\quad +\ours & \textcolor{gray}{70.8} & \textcolor{gray}{50.4} & \textbf{64.3} & \textbf{46.7} & \textbf{60.4} & 30.5 \\
\midrule
InternVL3.5-8B$^\dagger$ & \textcolor{gray}{66.7} & \textcolor{gray}{44.3} & 40.7 & 20.5 & 45.2 & 24.0 \\
\rowcolor{green!10}
\quad +\ours & \textcolor{gray}{72.4} & \textcolor{gray}{49.5} & \textbf{48.7} & \textbf{30.8} & \textbf{55.3} & \textbf{29.6} \\
\bottomrule
\end{tabular}
}
\label{tab:other_backbones}

\vspace{-1.5em}
\end{wraptable}

%% file: figure/fig_additional_results.tex
\begin{figure}[t]
\centering
\includegraphics[width=\linewidth]{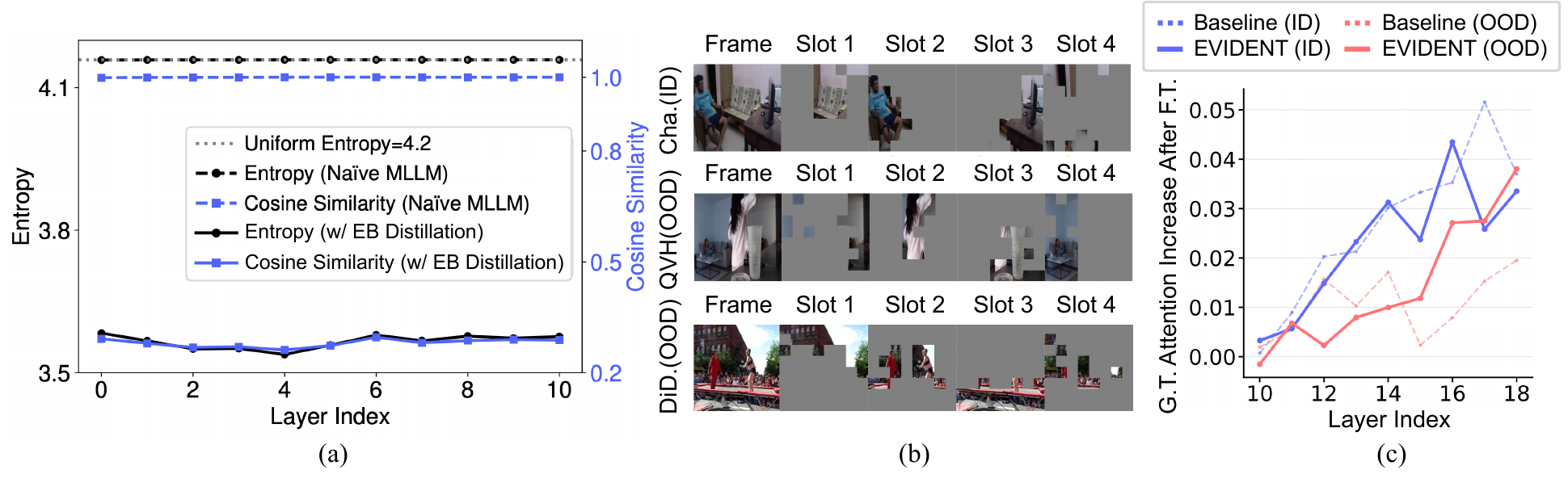}
\vspace{-2em}
\caption{
\textbf{Additional analysis on \ours{}.} 
(a) Entropy and pairwise cosine similarity of normalized slot attention weights in \adapter{} across layers.
(b) Slot attention maps of \ours{} on both ID and OOD samples, visualized by masking each frame with its highest-attending slot.
(c) Attention increase on G.T. frames relative to all frames after fine-tuning, compared between the baseline and \ours{}.
}
\label{fig:analysis}
\vspace{-1em}
\end{figure}

%% file: 6_conclusion.tex
\section{Conclusions}
\label{sec:conclusions}

We presented \textbf{EVIDENT}, a parameter-efficient framework for cross-domain Video Temporal Grounding that anchors temporal localization in the inherent entity-attention of pre-trained MLLMs. 
Our analysis identifies \emph{attention-localization decoupling} as a key failure in OOD VTG, which is primarily driven by visual domain shift rather than unseen query concepts. 
EVIDENT addresses this through three coupled components: an EB Adapter for routing visual tokens into entity-level slots, an EB Distillation loss for coherent slot bindings, and an E2V gating mechanism for entity-attention–based evidence grounding.
Experiments on cross-domain VTG benchmarks show that EVIDENT consistently improves OOD robustness while maintaining competitive ID performance. 
Our results highlight that evidence-grounded entity-attentive adaptation, rather than direct fitting to source-domain context, is the key to robust cross-domain VTG.

%% file: 9_supple.tex
\appendix

\renewcommand{\theHsection}{appendix.\Alph{section}}
\phantomsection
\section*{\Large \bf Appendix}

\setcounter{page}{1}
\setcounter{table}{0}
\setcounter{figure}{0}
\pagenumbering{roman}
\renewcommand{\thetable}{\Roman{table}}
\renewcommand{\thefigure}{\Roman{figure}}

In this appendix, we provide comprehensive analyses, method details, and additional quantitative results to complement the main paper. We organize the appendix as follows:

\begin{itemize}
    \item \textbf{\secref{supple:extended_rw}: Extended Related Work}
    \item \textbf{\secref{supp:analysis}: Further Analysis of the Domain Gap}
    \begin{itemize}
        \item \secref{supp:concept_split}: Constructing Seen/Unseen Concept Splits
        \item \secref{supp:visual_similarity}: Measuring Cross-Domain Visual Similarity
        \item \secref{supp:internvl}: Diagnosing on a Different MLLM
    \end{itemize}
    \item \textbf{\secref{supp:method}: Further Method Details}
    \begin{itemize}
        \item \secref{supp_concept_extraction}: Extracting Subject-Object Concepts for \gating{}
        \item \secref{supp:dino_viz}: Visualizing DINOv2 Cluster Maps as Supervision Signal
        \item \secref{imple_detail}: Implementation Details
    \end{itemize}
    \item \textbf{\secref{supp:experiments}: Additional Experimental Results}
    \begin{itemize}
        \item \secref{supp:ablation}: Additional Ablation Study
        \item \secref{supp:slot_viz}: Qualitative Analysis of \ours{}
    \end{itemize}
    \item \textbf{\secref{supp:limitations}: Limitations}
    \item \textbf{\secref{supp:broader_impact}: Broader Impact}
\end{itemize}

\section{Extended Related Work}
\label{sec:supple:extended_rw}
\paragraph{DETR-based VTG Specialists}
Prior VTG works follow the DETR paradigm~\cite{carion2020end} dominantly, where Moment-DETR~\cite{lei2021detecting} first cast moment retrieval as a set prediction problem and established the QVHighlights~\cite{lei2021detecting} benchmark.
Subsequent models extend this paradigm by injecting VTG-specific inductive biases into the decoder, such as query-dependent representations~\cite{moon2023query}, event-aware attention~\cite{jang2023knowing}, and correlation-guided calibration~\cite{moon2023correlation}.
More recently, MLLM-based VTG methods~\cite{wang2024hawkeye,zeng2025timesuite,li2025unitime,liu2026videomind,meinardus2024chrono} have emerged as a compelling alternative, outperforming DETR-based specialists.
In this work, we show that they still struggle to generalize to unseen domains, identify the visual domain gap as the primary cause, and address it with \ours{}.

\paragraph{Bias in Video Understanding}
Dataset bias has been widely studied in video understanding.
In action recognition, prior works have shown that models exploit static appearance cues, such as scenes and objects, as shortcuts~\cite{choi2019can,li2018resound,bae2024devias,ahn2026rcore}.
In multimodal video understanding, models have similarly been found to over-rely on background context or a single frame~\cite{lei2023revealing,bae2025mashvlm}.

In VTG, \citet{otani2020uncovering} show that blind baselines without video input can rival trained models by exploiting the moment location distribution of annotations.
This finding motivated benchmarks for evaluating generalization, including OOD re-splits~\cite{yuan2021closer}, compositional splits with novel query compositions~\cite{li2022compositional}, and a comprehensive benchmark across multiple datasets~\cite{chae2024towards}.
To mitigate temporal location bias, prior works apply causal intervention~\cite{yang2021deconfounded,nan2021interventional}, perturb or augment the temporal structure of videos~\cite{hao2022can,lan2023curriculum}, or synthesize bias-conflict samples~\cite{qi2024bssard}.
Others address cross-scenario transfer via sample reweighting~\cite{bao2022debiasttl} or query prompting~\cite{jiang2024mgqp}.
However, these methods mainly target non-MLLM grounding models, whereas we study the visual domain shortcuts that arise when fine-tuning MLLMs, and address them by preserving the pre-trained entity-attention rather than by explicit debiasing.

\section{Further Analysis of the Domain Gap}
\label{sec:supp:analysis}

\subsection{Constructing Seen/Unseen Concept Splits}
\label{sec:supp:concept_split}      

For the concept domain gap analysis shown in \figref{diagnosis}~(a), we partition the QVHighlights (QVH)~\cite{lei2021detecting} test set into \emph{seen} and \emph{unseen} concept subsets with respect to the Charades-STA (Cha.)~\cite{gao2017tall} training distribution. Specifically, we first build a Cha.-centric concept vocabulary $\mathcal{C}$ by taking the 50 most frequent content lemmas across all Cha. training queries, after standard lowercasing, lemmatization, and stop-word removal.  The resulting set covers the dominant indoor activity, object, and motion vocabulary of Cha. (e.g.\ \textit{glass, walk, sneeze, room, door, cup, shelf, sit, pillow, refrigerator, blanket}). Each QVH test query is then processed with the identical pipeline, and labeled \emph{seen} if at least one of its lemmas appears in $\mathcal{C}$, otherwise \emph{unseen}. This rule yields $635/915$ seen/unseen samples out of $1{,}550$; to remove the class-size imbalance, we draw a fixed random subset of 600 samples from each group as our final evaluation splits. We use only word-level lemma matching, without expanding to a hand-crafted concept ontology, so that the split reflects lexical familiarity to the Cha.-trained model without additional human bias.

\subsection{Measuring Cross-Domain Visual Similarity}
\label{sec:supp:visual_similarity}

\begin{figure}[htbp]
\centering
\includegraphics[width=\linewidth]{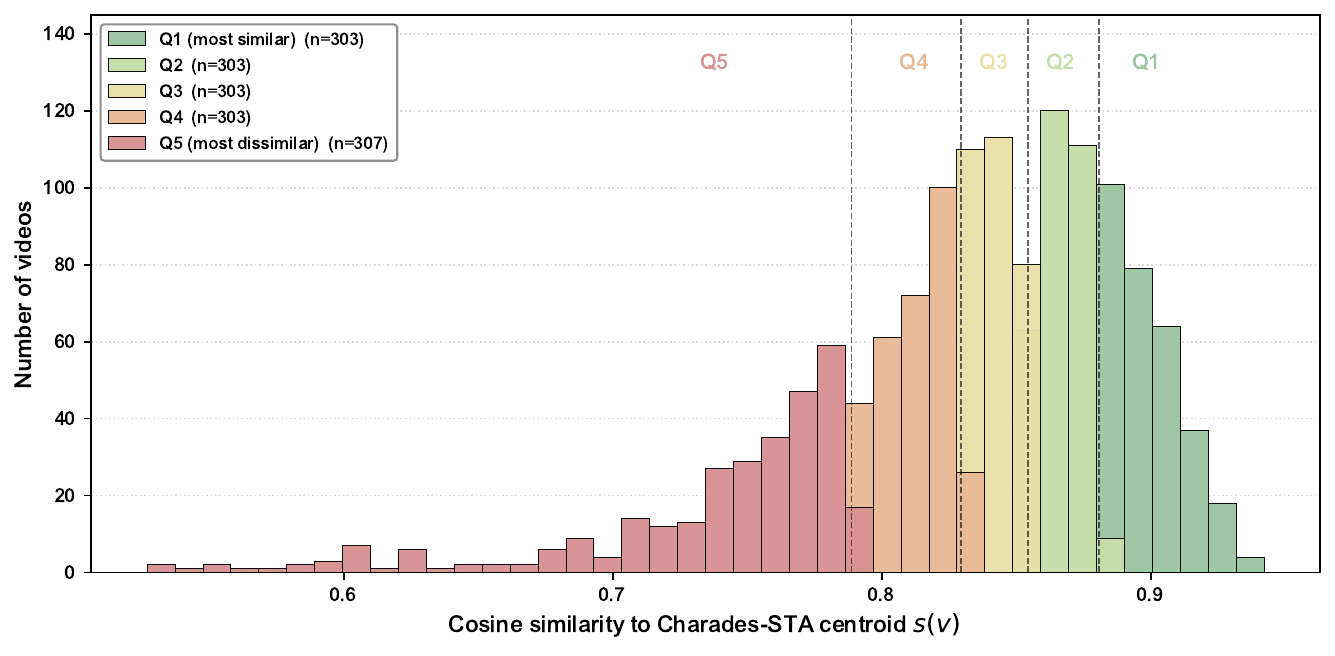}
\caption{
\textbf{Visual similarity distribution and quintile partitioning.} 
Distribution of cosine similarities $s(v) = \cos(\mathbf{v}, \mathbf{c}_{\text{Cha.}})$ between QVHighlights~\cite{lei2021detecting} test video descriptors $\mathbf{v}$ and the Charades-STA~\cite{gao2017tall} training centroid $\mathbf{c}_{\text{Cha.}}$, computed in a frozen Qwen2.5-VL-3B~\cite{bai2025qwen25vl} embedding space. The 1,519 test videos (1,550 query samples) are sorted by $s(v)$ and partitioned into five equally sized quantile bins (dashed lines). We evaluate temporal grounding on the highest- and lowest-similarity bins, \textbf{Q1} (\emph{Visually similar}, $s \geq 0.881$) and \textbf{Q5} (\emph{Visually dissimilar}, $s \leq 0.789$), to isolate the effect of visual domain shift. Cosine values are relative scores within this encoder's embedding space and are not directly comparable to similarities computed under a different encoder.
}
\label{fig:visual_bias_hist}
\end{figure}

For the visual domain gap analysis shown in \figref{diagnosis}~(b), we embed each video as a single global descriptor by passing 20 uniformly sampled $224{\times}224$ frames through a frozen Qwen2.5-VL-3B~\cite{bai2025qwen25vl} vision encoder, mean-pooling visual tokens within each frame and then averaging the resulting per-frame vectors. Taking Charades-STA (Cha.)~\cite{gao2017tall} training videos as the reference distribution, we define a centroid $\mathbf{c}_{\text{Cha.}}$ as the mean of their global descriptors, and characterize each test video by its cosine similarity $s(v) = \cos(\mathbf{v}, \mathbf{c}_{\text{Cha.}})$ to this centroid.
While the Cha. training set is internally tight (intra-set similarity $0.894 \pm 0.035$ over $5{,}335$ videos), QVHighlights (QVH)~\cite{lei2021detecting} test videos drift toward lower similarities ($0.831 \pm 0.061$, range $[0.527,\,0.942]$ over $1{,}519$ videos), indicating a measurable train$\rightarrow$test visual shift (Figure ~\ref{fig:visual_bias_hist}).

We sort the QVH test set by $s(v)$ and partition it into five equally sized quantile bins (Figure ~\ref{fig:visual_bias_hist}). Reporting performance on the top and bottom quantiles---denoted \textbf{Visually similar} (Q1, $s\!\geq\!0.881$) and \textbf{Visually dissimilar} (Q5, $s\!\leq\!0.789$)---isolates the impact of the visual domain gap on temporal grounding without conflating it with other dataset-specific factors. Note that the cosine values themselves are meaningful only as relative scores within the Qwen2.5-VL~\cite{bai2025qwen25vl} embedding space; we use them solely to induce an ordering over test videos.

\subsection{Diagnosing on a Different MLLM}
\label{sec:supp:internvl}
\input{figure/fig_internvl}

To verify that our problem analysis is not specific to Qwen2.5-VL~\cite{bai2025qwen25vl}, we repeat the analyses of \figref{observation} and \figref{diagnosis} on a different MLLM, InternVL3-2B~\cite{zhu2025internvl3}. As shown in \figref{internvl}, InternVL3-2B exhibits the same pattern as Qwen2.5-VL reported in \secref{background}.

\noindent\textbf{Entity-attention analysis (\figref{internvl}~(a)).}
The zero-shot InternVL3-2B exhibits strong visual-text alignment across both Charades-STA~\cite{gao2017tall} and QVHighlights~\cite{lei2021detecting}, but na\"ive fine-tuning suffers severe ID-OOD performance gaps. The fine-tuned model's attention on ground-truth-interval frames substantially increases on the trained domain (ID) but only marginally on the unseen domain (OOD), indicating that the learned visual-text alignment ability fails to transfer to other domains.

\noindent\textbf{OOD gap breakdown (\figref{internvl}~(b)).}
Splitting the OOD test set by query-concept overlap with ID yields only a marginal performance gap, while splitting by visual feature similarity to ID yields a much larger gap. This identifies the visual domain gap as the primary cause of OOD degradation, consistent with our observations on Qwen2.5-VL. Together, these results suggest that entity-attention bypass under visual domain shift is a general phenomenon across MLLMs rather than a model-specific artifact.

\section{Further Method Details}
\label{sec:supp:method}

\subsection{Extracting Subject-Object Concepts}
\label{sec:supp_concept_extraction}
Entity-to-eVidence (\gating, Section~\ref{sec:frame_gating}) takes a subject-object concept pair as textual input. For each training query, we extract this pair by prompting {Qwen3-4B-Instruct-2507}~\cite{yang2025qwen3} in a zero-shot (Figure~\ref{fig:concept-prompt}) to return a JSON object with two entries: a \texttt{Subject:} (primary actor) and an \texttt{Object:} (tangible interaction target), each restricted to a 1--2 word noun. Given the extracted concept words, we identify their token positions in the original query token sequence as $i_{\text{sub}}$ and $i_{\text{obj}}$, which are then used to select the corresponding query token features $\mathbf{q}'_{\text{sub}}$ and $\mathbf{q}'_{\text{obj}}$ for \gating. This schema deliberately excludes verbs, scenes, and abstract descriptions so that the resulting concepts correspond to \emph{visually grounded} entities that align with frame-level evidence during \gating.

\definecolor{promptvar}{RGB}{0,90,200}

\newtcolorbox{promptbox}{
    enhanced,
    colback=white,
    colframe=black!70,
    boxrule=0.6pt,
    arc=3mm,
    left=3mm, right=3mm, top=2mm, bottom=2mm,
    fonttitle=\bfseries,
}

\begin{figure}[h]
\centering
\begin{promptbox}
\textbf{\large User}\\[2pt]
You are an expert in semantic decomposition for video temporal grounding.
Extract the most important Subject and Object from the query for
frame-level grounding.\\[4pt]
\textbf{Roles \& Prefixes you MUST use:}
\begin{itemize}\itemsep0pt
    \item \texttt{"Subject:[noun]"} $\rightarrow$ The primary living
          actor or main entity.
    \item \texttt{"Object:[physical noun]"} $\rightarrow$ The primary
          tangible target object being interacted with. If none exists,
          omit this.
\end{itemize}
\textbf{Rules:}
\begin{enumerate}\itemsep0pt
    \item Return a JSON list containing up to 2 strings.
    \item Each string MUST start with \texttt{"Subject:"} or
          \texttt{"Object:"}.
    \item If the Subject is interacting with a physical item, you MUST
          extract it as an Object.
    \item Ignore verbs, scenes, abstract concepts, and meta-descriptions.
    \item Keep the nouns extremely concise (1--2 words).
    \item Return ONLY a valid JSON list.
\end{enumerate}
\textbf{Examples:}\\
Query: \textcolor{promptvar}{``a person opens the refrigerator in the kitchen''}\\
JSON list: \texttt{["Subject: person", "Object: refrigerator"]}\\[2pt]
Query: \textcolor{promptvar}{``The girl dances around the room.''}\\
JSON list: \texttt{["Subject: girl"]}\\[4pt]
Query: \textcolor{promptvar}{\{query\}}\\
JSON list:
\par\vspace{4pt}\hrule\vspace{4pt}
\textbf{\large Assistant}\\[2pt]
\textcolor{promptvar}{JSON list of extracted concepts}
\end{promptbox}
\caption{Zero-shot prompt used with Qwen3-4B-Instruct-2507~\cite{yang2025qwen3} for key
concept extraction. \textcolor{promptvar}{Blue} text represents
variables.}
\label{fig:concept-prompt}
\end{figure}

\subsection{Visualizing DINOv2 Cluster Maps as Supervision Signal}
\label{sec:supp:dino_viz}

Our \dinoloss{} relies on DINOv2~\cite{oquab2023dinov2} $K$-means cluster maps as pseudo-ground-truth supervision for slot attention. To verify the reliability of this supervision signal, we examine whether DINOv2 features produce semantically meaningful clusters.

\begin{figure}[h]
\centering
\includegraphics[width=\linewidth]{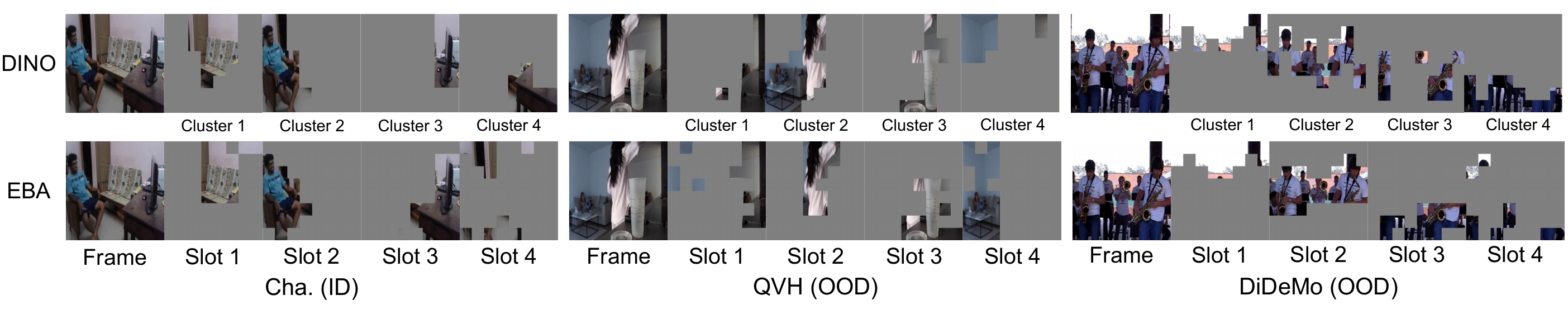}
\caption{
\textbf{DINOv2 cluster maps vs.\ \adapter{} slot assignments.}
Each group shows a frame and four isolated patch views across Charades-STA (Cha.)~\cite{gao2017tall}\ (ID), QVHighlights (QVH)~\cite{lei2021detecting}(OOD), and DiDeMo~\cite{anne2017localizing} (OOD).
\textbf{Top (DINO):} DINOv2~\cite{oquab2023dinov2} $K$-means naturally separates semantically coherent regions, validating its use as pseudo-ground-truth in \dinoloss{}.
\textbf{Bottom (\adapter{}):} Slots trained with \dinoloss{} mirror the DINO structure, binding each slot to a distinct entity consistently across domains.
}
\label{fig:dino_cluster_viz}
\end{figure}

We visualize DINOv2 cluster maps on videos from our evaluation benchmarks. Concretely, we extract DINOv2-large patch features for 20 uniformly sampled $224{\times}224$ frames, apply $2{\times}2$ average pooling to reduce each frame to an $8{\times}8$ spatial grid (64 patches), $\ell_2$-normalize the features, and run $K$-means with $K{=}4$. Figure~\ref{fig:dino_cluster_viz} shows the resulting cluster assignments: for each cluster, we display only the patches assigned to that cluster.

The results demonstrate that DINOv2 $K$-means clusters naturally partition frames into semantically coherent regions---consistently separating foreground actors, interaction targets, background scene elements, and motion-salient regions. This confirms that using DINOv2 cluster maps as pseudo-GT for \dinoloss{} provides a semantically grounded training signal rather than an arbitrary one, justifying our design choice. As shown in the bottom row of Figure~\ref{fig:dino_cluster_viz}, slots trained with our \dinoloss{} mirror this DINO structure, binding each slot to a distinct entity consistently across both ID and OOD domains.

\subsection{Implementation Details}
\label{sec:imple_detail}
We build upon Qwen2.5-VL-Instruct~\cite{bai2025qwen25vl} (7B) as the backbone MLLM, where the LLM decoder has a hidden dimension of $D{=}3584$. The vision encoder processes each $224{\times}224$ frame into $N{=}64$ visual tokens ($8{\times}8$ spatial grid).
For the \adapter{}, we set the bottleneck dimension to $d{=}896$, the number of slots to $K{=}4$, and use 8 attention heads with $I{=}3$ iterations of GRU-based refinement.
We inject the \adapter{} into layers 0--10, while LoRA~\cite{hu2022lora} (rank 16, $\alpha{=}64$) is applied to the remaining deeper layers; the vision tower, the LLM's frozen layers, and the vision--language merger are kept fixed throughout training.
The \gating{} module is applied at every \adapter{} layer (0--10) and modulates the slot residual using the cosine similarity between subject/object queries and per-frame mean-pooled visual tokens.
The \dinoloss{} derives pseudo-GT cluster maps by $K$-means clustering DINOv2-large~\cite{oquab2023dinov2} patch features, and aligns slot-attention maps to these clusters via Hungarian matching with BCE loss at every \adapter{} layer (0--10) with $\lambda{=}0.1$.
\paragraph{Video sampling and timestamps.}
We uniformly sample 20 and 60 frames for the models trained on Charades-STA~\cite{gao2017tall} and QVHighlights~\cite{lei2021detecting}, respectively. For cross-domain evaluation, we follow each \emph{target} dataset's native protocol and sample 20 frames for Charades-STA, 60 frames for QVHighlights, and 20 frames for DiDeMo~\cite{anne2017localizing}.
For Charades-STA training, whose clips are confined to a narrow $0$--$30$\,s range, we further adopt the \emph{relative-integer} timestamp encoding of Chrono-BLIP~\cite{meinardus2024chrono}, representing each timestamp as an integer in $[0,100]$ relative to clip duration; this mitigates the text bias toward a narrow band of numeric tokens that the dataset's limited duration range would otherwise induce.
\paragraph{Gating under partial concept queries.}
Not every query specifies both a subject and an object: queries such as \emph{``a person is dancing''} contain a subject but no
object. For such samples, we pass the absent concept's score through as $1$ rather than $0$: since the per-frame gate is the
multiplicative product of the two scores, $1$ is its identity element, so the absent concept exerts no influence on gating while
the present one continues to drive it---whereas $0$ would collapse the gate regardless of the present concept's signal.
\input{table/hyperparams}

\paragraph{Optimization.}
We train for 2 epochs with AdamW~\cite{adamw} (learning rate $5{\times}10^{-5}$, weight decay $0.1$) under a cosine schedule with 5\% linear warmup, using a global batch size of 32 on 8 NVIDIA 3090/4090 GPUs. Full hyperparameter details are summarized in \tabref{hyperparams}.
{In total, the trainable parameters amount to approximately 201M (2.42\% of the backbone) for the 7B model.}

\section{Additional Experimental Results}
\label{sec:supp:experiments}

\subsection{Additional Ablation Study}
\label{sec:supp:ablation}

We provide further ablations on the design choices of \ours{} in \tabref{ablation_supp}.

\input{table/supple_ablation}

\noindent\textbf{Effects of \adapter{} insertion layers.}
In \tabref{ablation_supp} (a), we show that inserting the \adapter{} into decoder layers 0–10 yields the best OOD performance. 
As discussed in \secref{slot_adapter}, this choice places our \adapter{} in the early layers where cross-frame interactions occur~\citep{maptheflow}, allowing each slot to capture temporally coherent disentangled semantics. 
Meanwhile, the mid-to-later layers responsible for vision-language integration~\citep{maptheflow} are left to lightweight LoRA~\citep{hu2022lora} fine-tuning, preventing excessive overfitting to the source domain.

\noindent\textbf{Effects of token reconstruction design.}
In \tabref{ablation_supp} (b), we demonstrate that reusing the slot attention weights to reconstruct the original token sequence, mentioned in \secref{slot_adapter}, substantially improves OOD performance (+2.1 points in R1@0.5). 
This attention reuse design routes VTG adaptation through disentangled entity-level representations and eliminates the additional learnable parameters that could overfit to the source domain.

\noindent\textbf{Effects of the number of slots.}
In \tabref{ablation_supp} (c), ID performance remains stable across $N_s$, whereas OOD performance degrades as $N_s$ increases (-5.9 R1@0.5 from $N_s{=}4$ to $16$).
We find that four slots are sufficient for entity-level binding, and additional slots are likely to split each entity into fragmented parts, weakening cross-domain transferability.

\noindent\textbf{Effects of visual feature source in \gating{}.}
In \tabref{ablation_supp} (d), we ablate which visual feature is used to compute the per-frame gating score, comparing tokens \emph{before} down projection, \emph{after} down projection, and \emph{after} slot-based reconstruction. Features \emph{after} down projection give the best overall performance, achieving the highest OOD scores (R1@0.5 of 59.8) while remaining competitive on ID. Pre-projection features underperform on both splits, and post-reconstruction features collapse on OOD (R1@0.5 drops to 51.9), as the slot bottleneck discards fine-grained details needed for entity-level scoring. This indicates that \gating{} should be computed from the LLM's visual stream rather than the bottlenecked slot output.

\noindent\textbf{Effects of \dinoloss{} loss scale and placement.}
In \tabref{ablation_supp} (e), we ablate the loss weight $\lambda$ and the layer indices at which \dinoloss{} is applied. Across the three loss scales tested under the all-layer setting (0--10), the differences are minor: ID and OOD metrics remain within roughly $1$--$2$ points across $\lambda \in \{1.0, 0.5, 0.1\}$, indicating that \dinoloss{} is robust to the choice of loss scale. Our default $\lambda{=}0.1$ achieves the best ID R1@0.5 ($69.1$) and OOD R1@0.7 ($36.9$), while $\lambda{=}1.0$ slightly favors OOD R1@0.5 ($60.4$ vs.\ $59.8$). Restricting \dinoloss{} to only the last \adapter{} layer ($l{=}10$) leads to a small but consistent drop across metrics, suggesting that progressively applying \dinoloss{} across all \adapter{} layers (0--10) is mildly more effective than aligning only the final layer.

\noindent\textbf{Effects of \gating{} applied layers.}
In \tabref{ablation_supp} (f), we ablate whether \gating{} should be applied at every \adapter{} layer (L0--10) or only at the final layer (L10). Applying \gating{} across all \adapter{} layers consistently outperforms the last-layer-only variant on OOD R1@0.5 ($59.8$ vs.\ $55.8$) while maintaining competitive ID performance. This shows that progressively gating the slot residual at every layer better aligns the entity-level representations with query-relevant frames, leading to more robust cross-domain generalization.

\noindent\textbf{Effects of attention normalization.}
\label{sec:supp:token_norm}
In \tabref{ablation_supp} (g), we compare two attention maps for token reconstruction (eq. \eqref{eq:attn_transpose}) and EB Distillation (eq. \eqref{eq:EBD_loss}): the slot-normalized map $\mathbf{A}$ and the token-normalized map $\hat{\mathbf{A}}$ used in the slot update.
While both variants perform comparably in the ID setting, $\hat{\mathbf{A}}$ yields higher OOD performance (+3.3 R1@0.5).

\subsection{Qualitative Analysis of \ours{}}
\label{sec:supp:slot_viz}

\begin{figure}[t]
\centering
\includegraphics[width=\linewidth]{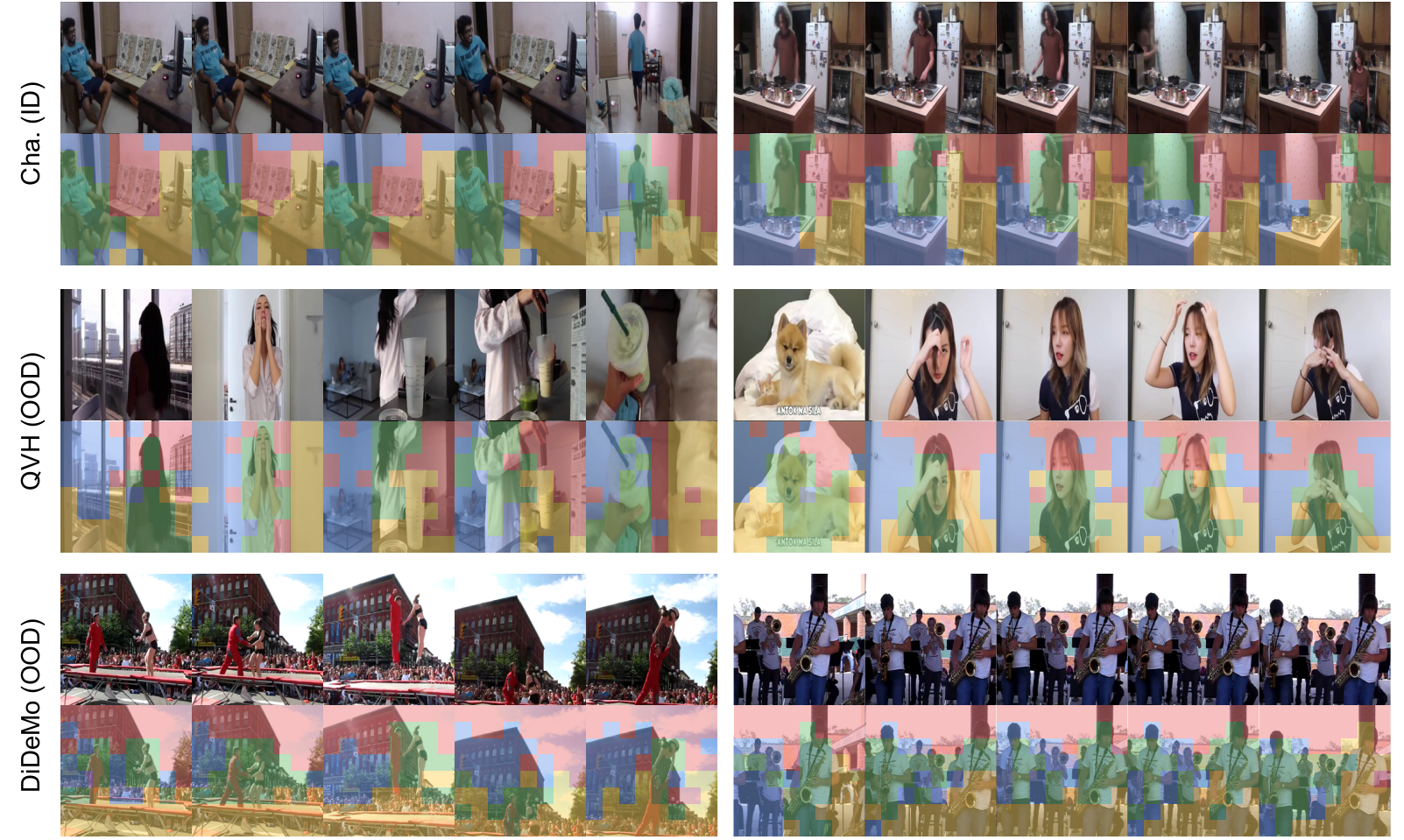}
  \caption{\textbf{\adapter{} visualization.} We visualize the slot assignments on samples from Charades-STA (Cha.)~\cite{gao2017tall}\ (ID), QVHighlights (QVH)~\cite{lei2021detecting} (OOD), and DiDeMo~\cite{anne2017localizing} (OOD) by masking each frame with its highest-attending slot. Frames are arranged in temporal order from left to right, and the same color denotes the same slot.}
\label{fig:slotviztime}
\end{figure}

\paragraph{Slot visualization in temporal order}
As shown in Figure~\ref{fig:slotviztime}, each slot consistently attends to the same entity across consecutive frames, forming temporally coherent trajectories along the time axis. Regions corresponding to the same slot (i.e., the same color) remain stable as the scene evolves, indicating that \adapter{} binds each slot to a specific entity and tracks it over time rather than re-assigning slots arbitrarily at each frame. This temporal continuity is observed not only on the in-domain Charades-STA~\cite{gao2017tall} samples but also on out-of-domain QVHighlights~\cite{lei2021detecting} and DiDeMo~\cite{anne2017localizing} samples, suggesting that the learned slot-entity binding generalizes beyond the training distribution.

\begin{figure}[t]
\centering
\includegraphics[width=.9\linewidth]{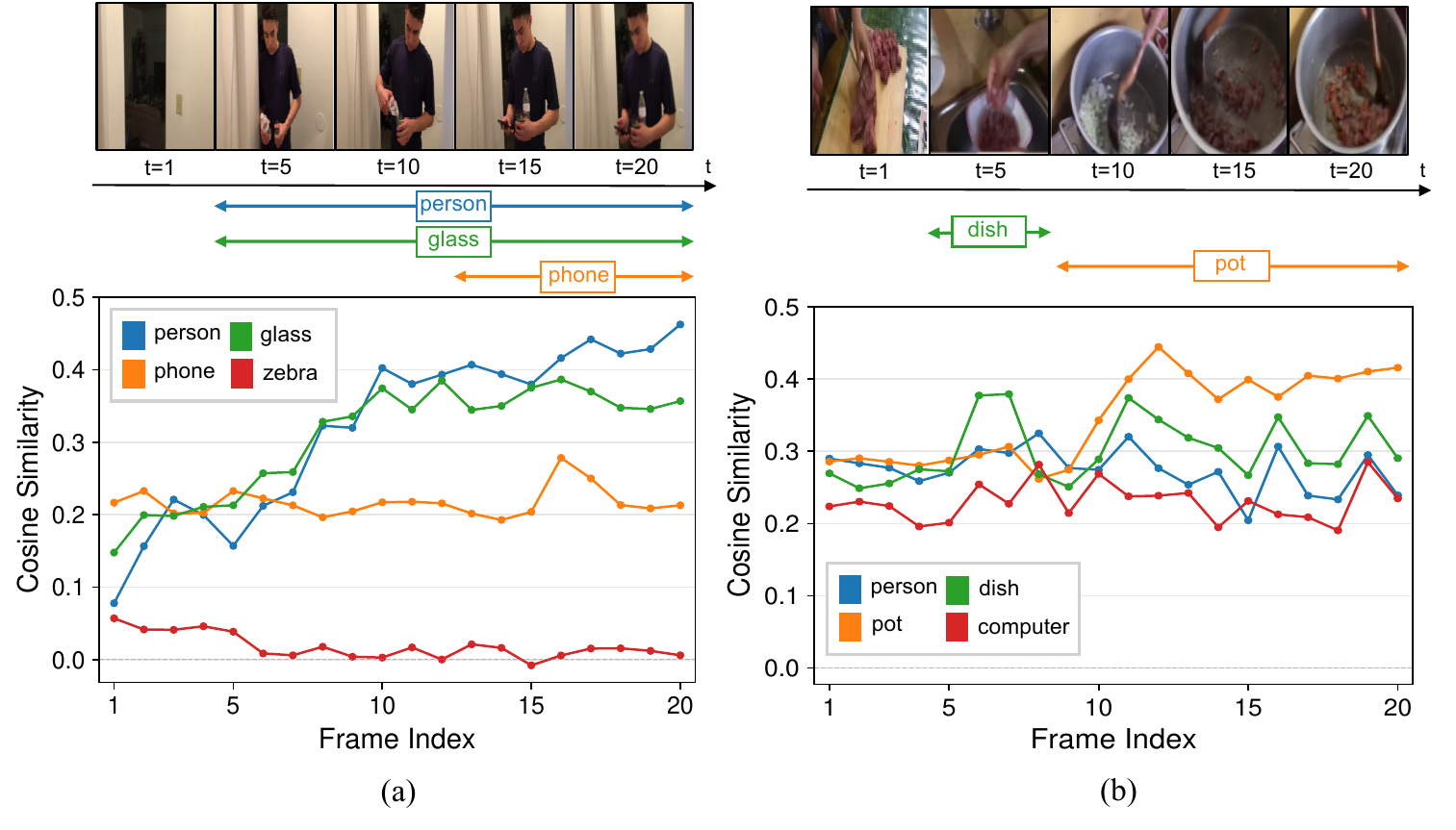}
\caption{\textbf{\gating{} score visualization.} 
We visualize the per-entity gating scores from \gating{} (see \secref{frame_gating}) on (a) Charades-STA~\cite{gao2017tall} (ID) and (b) QVHighlights~\cite{lei2021detecting} (OOD) samples.
}
\label{fig:gating_viz}
\end{figure}

\paragraph{Gating score visualization}
In Figure~\ref{fig:gating_viz}, we visualize the gating scores of each entity produced by \gating{} on Charades-STA~\cite{gao2017tall} (ID) and QVHighlights~\cite{lei2021detecting} (OOD) samples. We observe that the gating scores computed by \ours{} accurately reflect the temporal intervals in which each entity appears in the visual content.

\section{Limitations and Future Work}
\label{sec:supp:limitations}

EVIDENT addresses cross-domain VTG under visual distribution shift, with experiments spanning three diverse benchmarks (Charades-STA, QVHighlights, DiDeMo). Extending our analysis to more challenging settings such as long-form videos, egocentric perspectives, or open-vocabulary temporal grounding is a promising direction for future work. While we identify visual domain shift as the dominant failure mode of na\"ively fine-tuned MLLMs, jointly addressing other forms of distribution shift, such as temporal granularity or query style, would require complementary advances. We view EVIDENT as a first step toward systematically diagnosing and mitigating distribution shifts in MLLM-based video temporal grounding.

\section{Broader Impact}
\label{sec:supp:broader_impact}

EVIDENT improves the robustness of MLLM-based video temporal grounding under domain shift, with potential positive impacts on applications such as video search, content accessibility (e.g., caption alignment for visually impaired users), and video understanding tools for education and media analysis.
By promoting entity-grounded reasoning rather than reliance on dataset-specific shortcuts, our framework may also reduce systematic biases inherited from training data, contributing to more reliable model behavior in unseen scenarios.
We do not foresee direct negative societal impacts from this work, as it builds on publicly available pre-trained models and datasets and does not enable new generative or surveillance capabilities. Standard considerations applicable to general video understanding research, such as privacy concerns when applying such systems to personal video content, should be observed by downstream users.

%% file: figure/fig_internvl.tex
\begin{figure}[t]
  \centering
  \begin{subfigure}[t]{0.62\linewidth}
    \includegraphics[width=\linewidth]{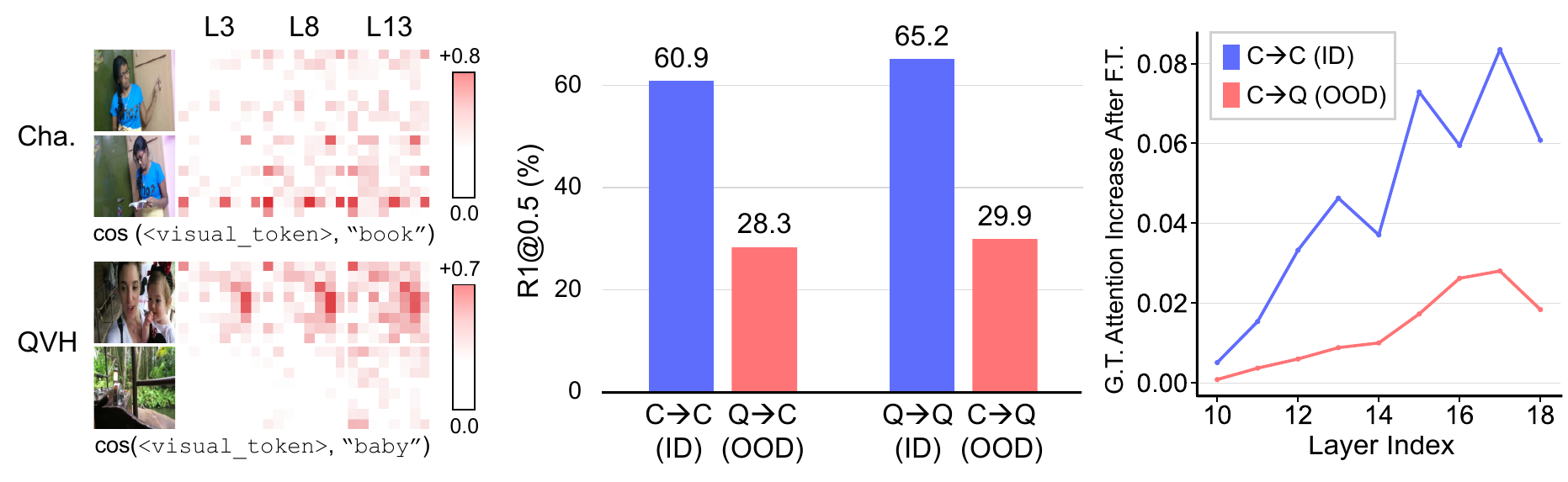}
    \caption{Entity-attention analysis}
    \label{fig:internvl_observation}
  \end{subfigure}
  \hfill
  \begin{subfigure}[t]{0.36\linewidth}
    \includegraphics[width=\linewidth]{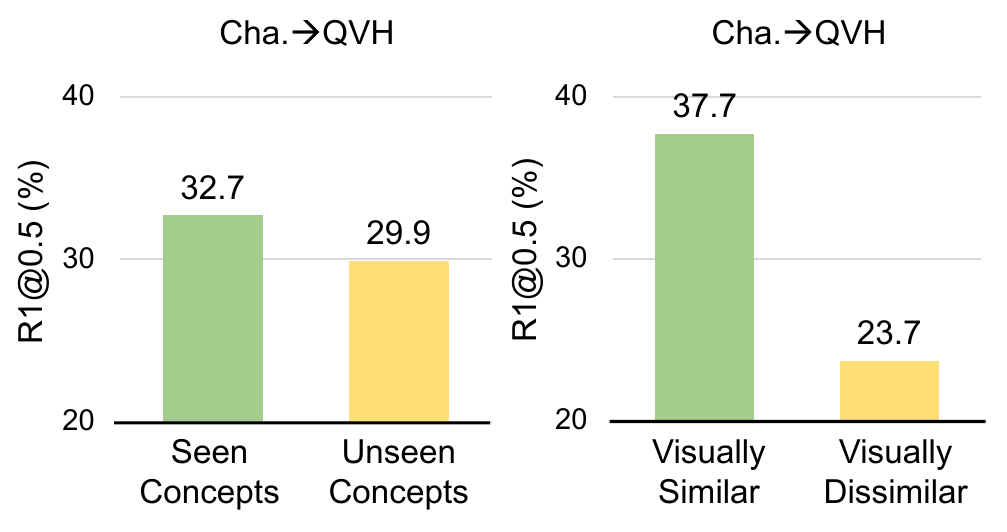}
    \caption{OOD gap breakdown}
    \label{fig:internvl_diagnosis}
  \end{subfigure}
  \caption{
    \textbf{Problem analysis on InternVL3-2B~\cite{zhu2025internvl3}.}
    We repeat the analysis of \figref{observation} and \figref{diagnosis} on a different MLLM, InternVL3-2B, and observe the same pattern as reported for Qwen2.5-VL~\cite{bai2025qwen25vl} in \secref{background}.
    \textbf{(a) Entity-attention analysis.} 
    (Left) Zero-shot entity-attention exhibits strong visual-text alignment regardless of the video domain. 
    (Middle) Na\"ive fine-tuning suffers severe ID-OOD gaps. 
    (Right) After fine-tuning, the attention on ground-truth-interval frames substantially increases on the trained domain (ID) but only marginally on the unseen domain (OOD), indicating that the learned visual-text alignment ability fails to transfer to other domains.
    \textbf{(b) OOD gap breakdown.}
    (Left) Splitting the OOD test set by query-concept overlap with ID yields only a marginal performance gap, indicating that the concept gap is not the dominant factor.
    (Right) Splitting OOD samples by visual feature similarity to ID yields a much larger gap, identifying the visual domain gap as the primary cause of OOD degradation.
  }
  \label{fig:internvl}
\end{figure}

%% file: table/hyperparams.tex
\begin{table}[t]
\centering
\small
\caption{\textbf{Fine-tuning setting of \ours{}}.}
\label{tab:hyperparams}
\begin{tabular}{@{}llcc@{}}
\toprule
Category & Hyperparameter & Charades-STA~\cite{gao2017tall} & QVHighlights~\cite{lei2021detecting} \\
\midrule
\multirow{2}{*}{Input}
  & \# Frames              & 20 & 60 \\
  & Resolution             & \multicolumn{2}{c}{$224 \times 224$} \\
\midrule
\multirow{6}{*}{Optimization}
  & Optimizer              & \multicolumn{2}{c}{AdamW~\cite{adamw}} \\
  & Learning rate          & \multicolumn{2}{c}{$5\times10^{-5}$} \\
  & LR scheduler           & \multicolumn{2}{c}{Cosine, 5\% warmup} \\
  & Weight decay           & \multicolumn{2}{c}{0.1} \\
  & Epochs                 & \multicolumn{2}{c}{2} \\
  & Global batch size      & \multicolumn{2}{c}{32} \\
\midrule  
\multirow{4}{*}{LoRA}
  & Applied layers         & \multicolumn{2}{c}{11--27} \\
  & Rank $r$               & \multicolumn{2}{c}{16} \\
  & $\alpha$               & \multicolumn{2}{c}{64} \\
  & Dropout                & \multicolumn{2}{c}{0.05} \\
\midrule
\multirow{5}{*}{\adapter{}}
  & Inserted layers        & \multicolumn{2}{c}{0--10} \\
  & \# Slots $K$           & \multicolumn{2}{c}{4} \\
  & Bottleneck dim         & \multicolumn{2}{c}{896} \\
  & \# Iterations          & \multicolumn{2}{c}{3} \\
  & Tokens per frame       & \multicolumn{2}{c}{64} \\
\midrule
\dinoloss{}
  & Loss weight $\lambda$  & 0.1 & 1.0 \\
\bottomrule
\end{tabular}
\end{table}

%% file: table/supple_ablation.tex
\begin{table*}[htbp]
\centering
\caption{\textbf{Additional Ablation Study.} 
Further ablations on the formulation of \dinoloss{} and the detailed configurations of \gating{}.
}

\mpage{0.48}{
(a) Effects of \adapter{} insertion layers.
\\
\resizebox{\linewidth}{!}{
\begin{tabular}{c c cc c cc}
\toprule
\multirow{2}{*}{Layer Index $i$} && \multicolumn{2}{c}{Cha. (ID)} & 
& \multicolumn{2}{c}{QVH (OOD)} \\ 
\cline{3-4} \cline{6-7} 
&& R1@0.5 & R1@0.7 && R1@0.5 & R1@0.7 \\
\midrule
0-6 && \textbf{69.2} & \textbf{49.5} && 56.8 & 34.9 \\
0-10 && 69.1 & 49.2 && \textbf{59.8} & \textbf{36.9} \\
0-14 && 69.1 & 49.4 && 58.3 & 33.7 \\
\bottomrule
\end{tabular}
}
}\hfill
\mpage{0.48}{
(b) Effects of token reconstruction design.
\\
\resizebox{\linewidth}{!}{
\begin{tabular}{l c cc c cc}
\toprule
\multirow{2}{*}{Method} && \multicolumn{2}{c}{Cha. (ID)} & 
& \multicolumn{2}{c}{QVH (OOD)} \\ 
\cline{3-4} \cline{6-7} 
&& R1@0.5 & R1@0.7 && R1@0.5 & R1@0.7 \\
\midrule
Cross attention && 69.0 & \textbf{49.3} && 57.7 & 35.1 \\
Attention reuse && \textbf{69.1} & 49.2 && \textbf{59.8} & \textbf{36.9} \\
\bottomrule
\end{tabular}
}
}

\mpage{0.48}{ 
{(c) Effects of the number of slots.}
\\
\resizebox{\linewidth}{!}{
\begin{tabular}{c c cc c cc}
\toprule
\multirow{2}{*}{\# of slots $N_s$} & \multicolumn{2}{c}{Cha. (ID)} & 
& \multicolumn{2}{c}{QVH (OOD)} \\ 
\cline{2-3} \cline{5-6} 
& R1@0.5 & R1@0.7 && R1@0.5 & R1@0.7 \\
\midrule
4 & 69.1 & 49.2 && \textbf{59.8} & \textbf{36.9} \\
8 & \textbf{69.2} & \textbf{49.4} && 56.6 & 36.6 \\
16 & \textbf{69.2} & 49.1 && 53.9 & 33.7
 \\
\bottomrule
\end{tabular}
}
}\hfill%
\mpage{0.5}{
(d) Effects of feature source in \gating{}. 
\\
\resizebox{\linewidth}{!}{
\begin{tabular}{l c cc c cc}
\toprule
\multirow{2}{*}{Visual feature} && \multicolumn{2}{c}{Cha. (ID)} & 
& \multicolumn{2}{c}{QVH (OOD)} \\ 
\cline{3-4} \cline{6-7} 
&& R1@0.5 & R1@0.7 && R1@0.5 & R1@0.7 \\
\midrule
Before down proj. && 68.4 & 49.0 && 55.8 & 34.1 \\ 
After down proj. && \textbf{69.1} & \textbf{49.2} && \textbf{59.8} & \textbf{36.9} \\
After token reconstruction && 69.2 & 50.3 && 51.9 & 29.7 \\ 
\bottomrule
\end{tabular}
}
}

\mpage{0.48}{ 
{\fontsize{9.5pt}{10.5pt}\selectfont 
(e) Effects of \dinoloss{} scale and placement.}
\\
\resizebox{\linewidth}{!}{
\begin{tabular}{c c cc c cc}
\toprule
\multirow{2}{*}{Loss scale $\lambda$} & \multirow{2}{*}{Layer index $l$} & \multicolumn{2}{c}{Cha. (ID)} & 
& \multicolumn{2}{c}{QVH (OOD)} \\ 
\cline{3-4} \cline{6-7} 
&& R1@0.5 & R1@0.7 && R1@0.5 & R1@0.7 \\
\midrule
1.0 & 0-10 & 68.3 & 49.2 && \textbf{60.4} & 36.0 \\
0.5 & 0-10 & 68.7 & \textbf{49.3} && 58.4 & 34.9 \\
0.1 & 0-10 & \textbf{69.1} & 49.2 && 59.8 & \textbf{36.9} \\
0.1 & 10 & 68.7 & 49.0 && 57.9 & 36.5 \\
\bottomrule
\end{tabular}
}
}\hfill
\mpage{0.48}{
(f) Effects of \gating{} applied layers. 
\\
\resizebox{\linewidth}{!}{
\begin{tabular}{c c cc c cc}
\toprule
\multirow{2}{*}{Layer Index $i$} && \multicolumn{2}{c}{Cha. (ID)} & 
& \multicolumn{2}{c}{QVH (OOD)} \\ 
\cline{3-4} \cline{6-7} 
&& R1@0.5 & R1@0.7 && R1@0.5 & R1@0.7 \\
\midrule
10 && 68.9 & \textbf{49.8} && 55.8 & 36.2 \\
0-10 && \textbf{69.1} & 49.2 && \textbf{59.8} & \textbf{36.9} \\ 
\bottomrule
\end{tabular}
}
}
\mpage{0.48}{
(g) Effects of attention normalization. 
\\
\resizebox{\linewidth}{!}{
\begin{tabular}{c c cc c cc}
\toprule
\multirow{2}{*}{Attention Map} && \multicolumn{2}{c}{Cha. (ID)} & 
& \multicolumn{2}{c}{QVH (OOD)} \\ 
\cline{3-4} \cline{6-7} 
&& R1@0.5 & R1@0.7 && R1@0.5 & R1@0.7 \\
\midrule
$\mathbf{A}$ && 68.7 & 48.9 && 56.5 & 33.4 \\
$\hat{\mathbf{A}}$ && \textbf{69.1} & \textbf{49.2} && \textbf{59.8} & \textbf{36.9} \\
\bottomrule
\end{tabular}
}
}
\label{tab:ablation_supp}
\end{table*}

%% file: checklist.tex
\section*{NeurIPS Paper Checklist}

\begin{enumerate}

\item {\bf Claims}
    \item[] Question: Do the main claims made in the abstract and introduction accurately reflect the paper's contributions and scope?
    \item[] Answer: \answerYes{}
    \item[] Justification: The abstract and \secref{intro} clearly state our central claim that robust cross-domain VTG requires evidence-grounded entity-attentive adaptation rather than direct fitting to source-domain context, and the contributions (problem analysis, EVIDENT framework, and cross-domain experiments) match the results reported in \secref{results}.


\item {\bf Limitations}
    \item[] Question: Does the paper discuss the limitations of the work performed by the authors?
    \item[] Answer: \answerYes{} %
    \item[] Justification: We discuss limitations and future directions in \secref{supp:limitations}, including the scope of cross-domain evaluation and other forms of distribution shift that complement our work.


\item {\bf Theory assumptions and proofs}
    \item[] Question: For each theoretical result, does the paper provide the full set of assumptions and a complete (and correct) proof?
    \item[] Answer: \answerNA{} %
    \item[] Justification: The paper presents an empirical study and does not include formal theoretical results requiring proofs.


    \item {\bf Experimental result reproducibility}
    \item[] Question: Does the paper fully disclose all the information needed to reproduce the main experimental results of the paper to the extent that it affects the main claims and/or conclusions of the paper (regardless of whether the code and data are provided or not)?
    \item[] Answer: \answerYes{} %
    \item[] Justification: We describe the architecture, training procedure, datasets, and evaluation protocol in \secref{method} and \secref{results}, and provide complete hyperparameters in \tabref{hyperparams} along with implementation details in \secref{imple_detail}.


\item {\bf Open access to data and code}
    \item[] Question: Does the paper provide open access to the data and code, with sufficient instructions to faithfully reproduce the main experimental results, as described in supplemental material?
    \item[] Answer: \answerNo{} %
    \item[] Justification: Code is not released at submission time to preserve anonymity, but we plan to release the code upon acceptance. All datasets used (Charades-STA, QVHighlights, DiDeMo) are publicly available.


\item {\bf Experimental setting/details}
    \item[] Question: Does the paper specify all the training and test details (e.g., data splits, hyperparameters, how they were chosen, type of optimizer) necessary to understand the results?
    \item[] Answer: \answerYes{} %
    \item[] Justification: Training and evaluation details, including data splits, hyperparameters, and optimization settings, are provided in \secref{results} and \secref{imple_detail}, with full hyperparameters listed in \tabref{hyperparams}.


\item {\bf Experiment statistical significance}
    \item[] Question: Does the paper report error bars suitably and correctly defined or other appropriate information about the statistical significance of the experiments?
    \item[] Answer: \answerNA{} %
    \item[] Justification: Following common practice in MLLM-based VTG~\cite{huang2024vtimellm, ren2024timechat, meinardus2024chrono, zeng2025timesuite, liu2026videomind}, we report results from a single training run due to the substantial computational cost of fine-tuning 7B-scale MLLMs. Our improvements over baselines are consistent across multiple datasets and source-target configurations (Table~\ref{tab:main}), suggesting that the gains are not due to seed variance.


\item {\bf Experiments compute resources}
    \item[] Question: For each experiment, does the paper provide sufficient information on the computer resources (type of compute workers, memory, time of execution) needed to reproduce the experiments?
    \item[] Answer: \answerYes{} %
    \item[] Justification: Compute resources (8 NVIDIA 3090/4090 GPUs) and training duration (2 epochs) are specified in \secref{imple_detail}.

    
\item {\bf Code of ethics}
    \item[] Question: Does the research conducted in the paper conform, in every respect, with the NeurIPS Code of Ethics \url{https://neurips.cc/public/EthicsGuidelines}?
    \item[] Answer: \answerYes{} %
    \item[] Justification: The research uses publicly available datasets, does not involve human subjects, and conforms with the NeurIPS Code of Ethics.


\item {\bf Broader impacts}
    \item[] Question: Does the paper discuss both potential positive societal impacts and negative societal impacts of the work performed?
    \item[] Answer: \answerYes{} %
    \item[] Justification: We discuss potential positive impacts (e.g., video accessibility, reduced bias in cross-domain settings) and considerations for downstream use in \secref{supp:broader_impact}.

    
\item {\bf Safeguards}
    \item[] Question: Does the paper describe safeguards that have been put in place for responsible release of data or models that have a high risk for misuse (e.g., pre-trained language models, image generators, or scraped datasets)?
    \item[] Answer: \answerNA{} %
    \item[] Justification: We do not release new pre-trained models or scraped datasets that pose a high risk for misuse. Our work builds on publicly available pre-trained MLLMs.


\item {\bf Licenses for existing assets}
    \item[] Question: Are the creators or original owners of assets (e.g., code, data, models), used in the paper, properly credited and are the license and terms of use explicitly mentioned and properly respected?
    \item[] Answer: \answerYes{} %
    \item[] Justification: We cite all datasets (Charades-STA, QVHighlights, DiDeMo) and pre-trained models (Qwen2.5-VL, InternVL3, DINOv2, Qwen3) used, and respect their respective licenses for academic research.


\item {\bf New assets}
    \item[] Question: Are new assets introduced in the paper well documented and is the documentation provided alongside the assets?
    \item[] Answer: \answerNA{} %
    \item[] Justification: This paper does not release new datasets or models at submission time.


\item {\bf Crowdsourcing and research with human subjects}
    \item[] Question: For crowdsourcing experiments and research with human subjects, does the paper include the full text of instructions given to participants and screenshots, if applicable, as well as details about compensation (if any)? 
    \item[] Answer: \answerNA{} %
    \item[] Justification: This research does not involve crowdsourcing or human subjects.


\item {\bf Institutional review board (IRB) approvals or equivalent for research with human subjects}
    \item[] Question: Does the paper describe potential risks incurred by study participants, whether such risks were disclosed to the subjects, and whether Institutional Review Board (IRB) approvals (or an equivalent approval/review based on the requirements of your country or institution) were obtained?
    \item[] Answer: \answerNA{} %
    \item[] Justification: This research does not involve human subjects and therefore does not require IRB approval.


\item {\bf Declaration of LLM usage}
    \item[] Question: Does the paper describe the usage of LLMs if it is an important, original, or non-standard component of the core methods in this research? Note that if the LLM is used only for writing, editing, or formatting purposes and does \emph{not} impact the core methodology, scientific rigor, or originality of the research, declaration is not required.
    \item[] Answer: \answerYes{} %
    \item[] Justification: We use Qwen3-4B-Instruct-2507~\cite{yang2025qwen3} as a preprocessing component to extract subject-object concept words from natural language queries, which is described in \secref{frame_gating} and detailed in \secref{supp_concept_extraction}.


\end{enumerate}